%% file: DistillationVAE.tex
\RequirePackage{fix-cm}
\documentclass[smallcondensed]{svjour3}     
\usepackage{graphicx}
\usepackage{booktabs}
\usepackage{natbib}
\usepackage{hyperref}
\usepackage[ruled,vlined]{algorithm2e}

\usepackage{subfigure}
\usepackage{microtype}
\usepackage{algorithmic}
\input{math_commands.tex}

\input{dis_command.tex}

\begin{document}

\title{DEFT: Distilling Entangled Factors by Preventing Information Diffusion\thanks{*Corresponding author}
}


\author{Jiantao Wu         \and
        Lin Wang*  \and
        Bo Yang   \and
        Fanqi Li \and
        Chunxiuzi Liu \and
        Jin Zhou
}


\institute{Jiantao Wu \and Lin Wang \and Bo Yang \and Fanqi Li \and Chunxiuzi Liu  \and Jin Zhou\at{} 
                Shandong Provincial Key Laboratory of Network Based Intelligent Computing, \\
                University of Jinan, Jinan 250022, China,  \\
              \email{wangplanet@gmail.com} (L. Wang)           
}

\date{Received: date / Accepted: date}

\maketitle

\begin{abstract}
  Disentanglement is a highly desirable property of representation owing to its similarity to human understanding and reasoning. Many works achieve disentanglement upon information bottlenecks (IB). Despite their elegant mathematical foundations, the IB branch usually exhibits lower performance. In order to provide an insight into the problem, we develop an annealing test to calculate the information freezing point (IFP), which is a transition state to freeze information into the latent variables. We also explore these clues or inductive biases for separating the entangled factors according to the differences in the IFP distributions. We found the existing approaches suffer from the information diffusion problem, according to which the increased information diffuses in all latent variables.

  Based on this insight, we propose a novel disentanglement framework, termed the distilling entangled factor (DEFT), to address the information diffusion problem by scaling backward information. DEFT applies a multistage training strategy, including multigroup encoders with different learning rates and piecewise disentanglement pressure, to disentangle the factors stage by stage. We evaluate DEFT on three variants of dSprite and SmallNORB, which show low-variance and high-level disentanglement scores. Furthermore, the experiment under the correlative factors shows incapable of TC-based approaches. DEFT also exhibits a competitive performance in the unsupervised setting.

\end{abstract}
\input{Body.tex}


%
%

\bibliographystyle{spbasic}      
\bibliography{disentanglement_simple}

\end{document}

%% file: math_commands.tex
\usepackage{amsmath,amsfonts,bm}

\def\eqref#1{equation~\ref{#1}}

\def\1{\bm{1}}

\def\vc{{\bm{c}}}

\DeclareMathAlphabet{\mathsfit}{\encodingdefault}{\sfdefault}{m}{sl}
\SetMathAlphabet{\mathsfit}{bold}{\encodingdefault}{\sfdefault}{bx}{n}

\newcommand{\E}{\mathbb{E}}

\DeclareMathOperator*{\argmax}{arg\,max}

%% file: dis_command.tex
\newcommand{\betavae}{$\beta$-VAE}

\newcommand{\btcvae}{$\beta$-TCVAE}


%% file: Body.tex
\section{Introduction}
An understanding and reasoning about the world based on a limited set of observations is important in the field of artificial intelligence.
For instance, we can infer the movement of a ball in motion at a single glance, as the human brain is capable of disentangling positions from a set of images without supervision.
Therefore, disentanglement learning is highly desirable to build intelligent applications.
A disentangled representation has been proposed to be beneficial for a large variety of downstream tasks \citep{Schlkopf2012OnCA,Peters2017ElementsOC}.
According to \citet{Kim.2018}, a disentangled representation promotes interpretable semantic information, resulting in substantial advancement, which includes but is not limited to reducing the performance gap between humans and AI approaches \citep{Lake.2017,DBLP:conf/iclr/HigginsSMPBBSBH18}.
Other instances of disentangled representations include semantic image understanding and generation \citep{Lample.2017}, zero-shot learning \citep{Zhu.2019}, and reinforcement learning \citep{HigginsPRMBPBBL17}.

As depicted in the seminal paper by \citet{Bengio.2013}, humans can understand and reason from a complex observation, after which they can reduce the explanatory factors. 
The observations are generated by explanatory ground-truth factors \(\vc\), which are invisible from the observations.
The task of disentanglement learning aims to obtain a disentangled representation that separates these factors from the observations.
The notion of disentanglement remains an open topic \citep{Higgins.2018,Do.2020}, and we follow a strict version of discourse that one and only one latent variable \(z_i\) represents one corresponding factor, \(c_j\) \citep{Burgess.2018}.

\citet{Locatello.2019} proved the impossibility of disentanglement learning without inductive biases on the model and data.
One popular inductive bias on the model assumes that the latent variables are independent.
These approaches, which are based on total correlation (TC), dominate visual disentanglement learning \citep{Chen.2018,D.S.Kim.2018,Kumar2018DIPVAE}.
This assumption is correct when the factors are sampled uniformly; however, the independent factors show statistical relevance in reality \citep{CorrelatedData.2021}.
For instance, we observe that men are more likely to have short hair, and based on the observations, there is a correlation between gender and hair length.
However, a man who is not bald may grow long hair if desired.
In other words, sex does not determine hair length, and they are two independent factors.
Therefore, the exploration of disentanglement approaches beyond the independence assumption is vital to reality applications.

Another popular research approach is based on information theory \citep{InsuJeon.2019,DBLP:conf/nips/ChenCDHSSA16}.
They hypothesize that the gradually increased information bottleneck (IB) leads to a better disentanglement \citep{Burgess.2018,Dupont.2018}.
Unfortunately, in practice, the approaches based on IB usually exhibit lower performance than those penalizing the TC \citep{Locatello.2019}.
However, it is important to understand whether this means that the total correlation beats the IB\@. It is believed that the answer is negative. 
In this research, we investigate the reason for which IBs fall behind TC in practice. \textit{We found that the information diffusion (ID) problem is an invisible hurdle that should be addressed in the IB community.}   

\begin{figure}[t]
    \centering
    \includegraphics[width=\linewidth]{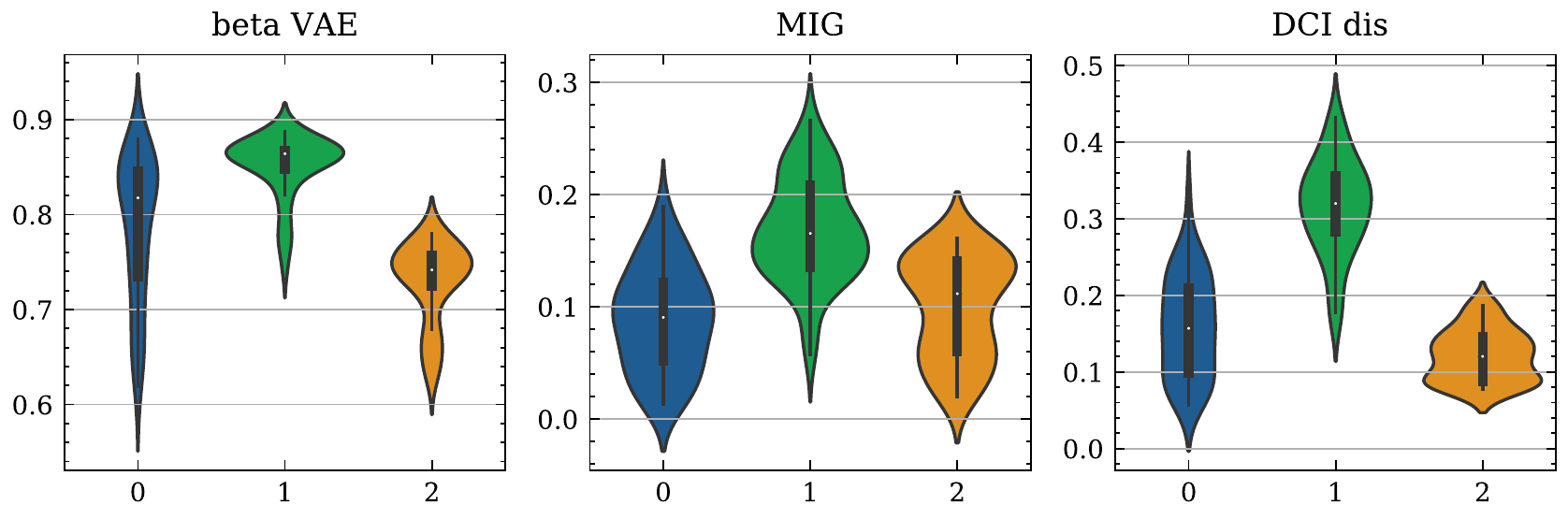}
    \caption{The distribution of beta VAE metric, MIG, and DCI disentanglement on dSprites. Models are abbreviated
        (0=\betavae{}, 1=\btcvae{}, 2=AnnealedVAE), and 50 trials are run with different random seeds.}%
    \label{fig:unstable_models}
\end{figure}

\textbf{Information diffusion} indicates that one factor’s information diffuses into two or more latent variables; thus, the disentanglement scores fluctuate during training.
Fig.~\ref{fig:unstable_models} shows the disentanglement scores of three approaches with the best hyperparameter settings, and it is observed that a large number of trials have a high variance\footnote{We use the pretrained models in \href{https://github.com/google-research/disentanglement\_lib}{disentanglement lib} by \citeauthor{Locatello.2019}.}.
We bridge the ID problem with the instability of the current approaches in Section~\ref{sec:fluctuation}.

In this paper, we trace the ID problem by measuring the NMI1.
The learned information may diffuse when AnnealedVAE and CascadeVAEC learn new information.
We have developed the annealing test to measure information freezing point (IFP) that the critical value for learning information from inputs.
The current IB-base approaches may prefer data with significant separable factors, because IB assumes the latent components have different contributions \citep{Burgess.2018}.

Inspired by distillation\footnote{Distillation is the process of separating a mixture into its components by heating at an appropriate temperature, such that components boil and freeze into the target containers.} in chemistry, we can divide the training process into several stages and extract one component at each stage.
In particular, we propose a framework, called the distilling entangled factor (DEFT), to disentangle factors stage-by-stage.
According to the IFP distribution, DEFT chooses selective pressure to enable some of the information to pass through the IB.
In addition, DEFT reduces the backward information of the first \(m\) encoders by reducing the learning rate to reduce the ID problem.
We evaluate DEFT on three four datasets, which shows robust performances.
We also exam DEFT on the dataset with correlative factors.
We have published our codes and all experimental settings in \href{https://gitee.com/microcloud/disentanglement_lib.git}{dlib for PyTorch} forked from \href{https://github.com/google-research/disentanglement_lib}{disentanglement lib}.
Our contributions are summarized in the following:
\begin{itemize}
    \item We hypothesize that the ID problem is one reason for the low performances of IB-based approaches.
    \item We propose DEFT, a multistage disentangling framework, to address the ID problem by scaling the backward information.
\end{itemize}

\section{Preliminary}

\subsection{Disentanglement Approaches}

\paragraph{Variational Autoencoder}
In variational inference, posterior \(p(z|x)\) is intractable.
The variational autoencoder (VAE) \citep{Kingma.2013} uses a neural network \(q_\phi(z|x)\) (encoder) to approximate the posterior \(p(z|x)\).
The other neural network \(p_\theta (x|z)\) (decoder) rebuilds the observations.
The objective of the VAE is to optimize the evidence lower bound (ELBO):
\begin{equation}\label{eq:vae}
    \begin{aligned}
        \mathcal{L}(\theta, \phi) =  \E_{q_\phi(\mathbf{z}|\mathbf{x})}[\log{p_\theta (x|z)}]
        - D_{\mathrm{KL}}(q_\phi(z|x) || p(z)).
    \end{aligned}
\end{equation}

\paragraph{\betavae}
\citeauthor{Higgins.2017} discovered the relationship between the disentanglement and the Kullback--Liebler (KL) divergence penalty strength.
They proposed the \(\beta\)-VAE to introduce additional pressure on the KL term:
\begin{equation}\label{eq:beta-vae}
    \begin{aligned}
        \mathcal{L}^1(\theta, \phi; \beta) =  \E_{q_\phi(\mathbf{z}|\mathbf{x})}[\log{p_\theta (x|z)}]
        - \beta D_{\mathrm{KL}}(q_\phi(z|x) || p(z)).
    \end{aligned}
\end{equation}
\(\beta \) controls the pressure for the posterior, \(q_{\phi}(z|x)\), to match the factorized unit Gaussian prior, \(p(z)\).
However, there is a trade-off between the quality of the reconstructed images and disentanglement.

\paragraph{AnnealedVAE}
\citet{Burgess.2018} proposed the AnnealedVAE, which progressively increases the information capacity of the latent variables while training:
\begin{equation}\label{eq:annealed-vae}
    \begin{aligned}
        \mathcal{L}^2(\theta, \phi; C) =  \E_{q_\phi(\mathbf{z}|\mathbf{x})}[\log{p_\theta (x|z)}]
        - \gamma \left| D_{\mathrm{KL}}(q_\phi(z|x) || p(z)) - C \right|,
    \end{aligned}
\end{equation}
where \(\gamma\) is a sufficiently large constant (usually 1000) to constrain the latent information, and \(C\) is a value that gradually increases from zero to a large number.

\paragraph{\btcvae}
The TC \cite{Watanabe.1960} quantifies the dependency among variables.
\btcvae{} \citep{Chen.2018} decomposed the KL term into three parts: mutual information (MI), total correlation (TC), and dimensional-wise KL (DWKL).
They proposed that the TC be penalized to achieve high reconstruction quality and disentanglement:
\begin{equation}
    \begin{aligned}
        \label{eq:btcvae}
        \mathcal{L}^3(\theta, \phi; \beta) = & \E_{q_\phi(\mathbf{z}|\mathbf{x})}[\log{p_\theta (x|z)}] -
        \mathbb{E}_{q(z, n)}\left[\log \frac{q_\phi(z \mid n) p(n)}{q_\phi(z) p(n)}\right] -                                                      \\
                                             & \beta \mathbb{E}_{q_\phi(z)}\left[\log \frac{q_\phi(z)}{\prod_{j} q_\phi\left(z_{j}\right)}\right]
        -\sum_{j} \mathbb{E}_{q_\phi\left(z_{j}\right)}\left[\log \frac{q_\phi\left(z_{j}\right)}{p\left(z_{j}\right)}\right].
    \end{aligned}
\end{equation}

\paragraph{CascadeVAEC}
\citeauthor{JeongS19a} provided another total correlation penalization through information cascading.
They proved that \(TC(z)=\sum_{i=2}^{d} I(z_{1:i-1};z_i)\).
CascadeVAEC, the continuous version, sequentially relieves one latent variable at a time, encouraging the model to disentangle one factor during the \(i\)-th stage:
\begin{equation}\label{eq:cascade-vae}
    \begin{aligned}
        \mathcal{L}^4(\theta, \phi; \beta_l,\beta_h) = & \E_{q_\phi(\mathbf{z}|\mathbf{x})}[\log{p_\theta (x|z)}] - \\
                                                       & \beta_l D_{\mathrm{KL}}(q_\phi(z_{1:i}|x) || p(z_{1:i})) -
        \beta_h D_{\mathrm{KL}}(q_\phi(z_{i+1:d}|x) || p(z_{i+1:d})),
    \end{aligned}
\end{equation}
where \(\beta_l\) is a small value for opening the information flow, \(\beta_h\) is a large beta coefficient, and \(d\) is the number of dimensions.

\subsection{Disentanglement Evaluation}

\paragraph{Supervised Disentanglement Metric}
Artificial datasets usually have ground-truth labels, such as dSprites \citep{dsprites17}.
The supervised metrics use this label information to measure the disentanglement responsibly.
Several metrics have been proposed to evaluate the disentanglement. These include the BetaVAE metric \citep{Higgins.2017}, FactorVAE metric \citep{Kim.2018}, MI gap \citep{Chen.2018}, modularity \citep{Ridgeway.2018}, DCI \citep{Eastwood.2018}, and SAP score \citep{Kumar2018DIPVAE}.
Shannon MI is an information-theoretic quantity that measures the amount of information shared between two variables.
The MIG measures the gap between the top two latent variables with the highest MI \citep{Chen.2018}.
\(j^1=\argmax_j I(z_j;c_k)\) is the index of latent variables that maximize the largest MI for \(c_k\). \(j^2=\argmax_j I(z_j;c_k), j \neq j^1\) is the second largest MI.
We define the \(m\)-th largest normalized MI (NMI) between \(z_j\) and \(c_k\) as \(\text{NMI}(c_k,m)\):
\begin{equation}
    \begin{aligned}
        \text{NMI}(c_k,m) = 
            \frac{1}{H\left(c_{k}\right)} I(z_{j^m};c_k).
    \end{aligned}
\end{equation}

\(\text{NMI}(c_k,2)\) is the largest \(\frac{I(z_j;c_k)}{H(c_k)} \) except \(\text{NMI}(c_k,1)\).
The MIG can be calculated as
\begin{equation}\label{eq:mig}
    \text{MIG} = \frac{1}{\Vert \vc \Vert} \sum_{i=1}^{\Vert \vc \Vert} \text{NMI}(c_i,1) - \text{NMI}(c_i,2).
\end{equation}

\section{Motivation}

\subsection{Disentanglement Fluctuation}\label{sec:fluctuation}

\begin{figure}
    \centering
    \includegraphics[width=\linewidth]{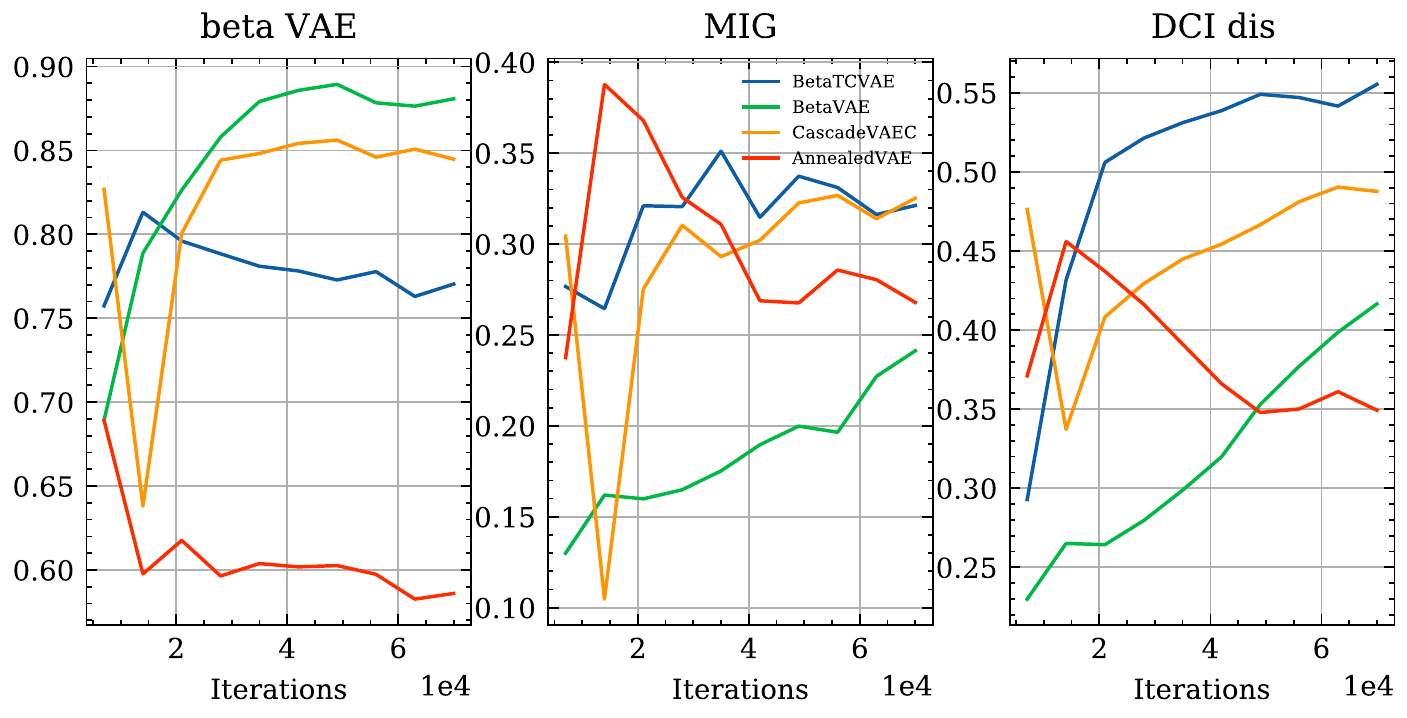}
    \caption{Disentanglement fluctuation for the IB-based approaches. AnnealedVAE and CascadeVAEC degeneracy into lower disentanglement scores.}%
    \label{fig:fluctuation}
\end{figure}
\citeauthor{Locatello.2019} conducted a survey of current disentanglement approaches, and the results show that these approaches have high variance of disentanglement scores. They concluded that ``tuning hyperparameters matters more than the choice of the objective function'' (See Figure 7 in their paper).
A reliable and robust approach should therefore have a consistently high performance and low variance.
We investigated the performance of \betavae{} (\(\beta=4\)), \btcvae{} (\(\beta=6\)), and AnnealedVAE (\(C=25\)) on dSprites, and traced the disentanglement scores through the training processes.
Fig.~\ref{fig:fluctuation} shows the curves of three metrics (beta VAE metric, MIG, and DCI disentanglement) for four models (\betavae{}, \btcvae{}, CascadeVAEC, and AnnealedVAE). 
AnnealedVAE, CascadeVAEC, and \btcvae{} show significant improvements in the very first iteration.
However, CascadeVAEC has a sharp decrement in the 1e4 iteration, and AnnealedVAE shows a downward trend after 1e4 iteration.
The training process did not consistently enhance the model being disentangled, resulting in poor performance.

\subsection{Information Bottleneck}
One solution to address fluctuation is to block some of the information by using a narrow bottleneck and then assign the increased information to a new latent variable by increasing the bottleneck.
AnnealedVAE and CascadeVAEC follow this concept; however, differ in terms of expanding the IB\@.
AnnealedVAE directly controls the capacity of the latent variables by an annealed increasing parameter, \(C\).
CascadeVAEC increases the capacity by relieving the pressure on the $i$-th latent variable at the $i$-th stage, opening the information flow.
Ideally, these approaches that are based on IB should have a steady growth of disentanglement; however, they also show fluctuation.

\subsection{Information Diffusion}\label{sec:diffusion}
\begin{figure}[t]
    \centering
    \subfigure[\(\text{NMI}(\vc,1)\)]{\includegraphics[width=.4945\linewidth]{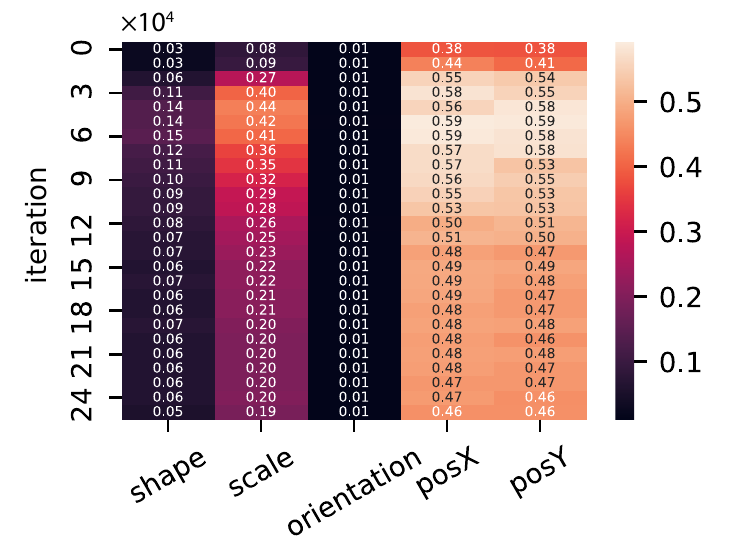}}%
    \subfigure[\(\text{NMI}(\vc,2)\)]{\includegraphics[width=.4945\linewidth]{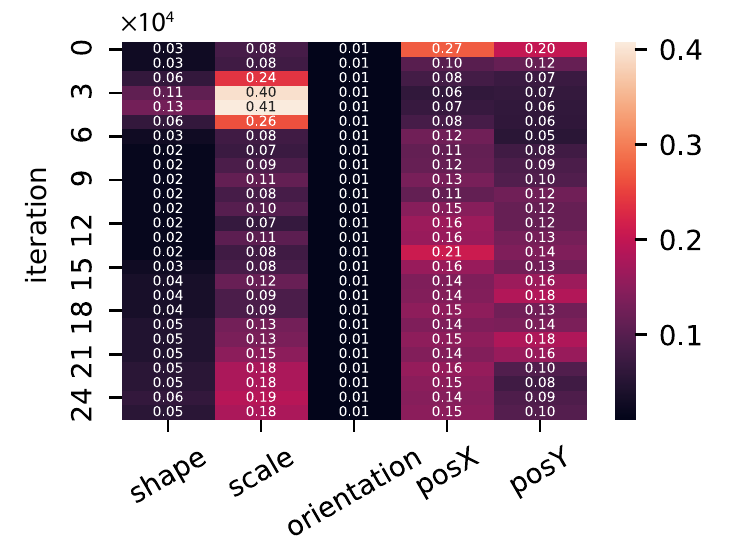}}
    \caption{The change of NMI over traning process on dSprites. The NMIs on many factors decrease gradually after the largest values have been captured at the early stage, especially on the factor scale.}\label{fig:NMI_itrs}
\end{figure}

A perfect disentangled representation should project one factor into one latent variable. 
In other words, the largest NMI, \(\text{NMI}(\vc,1)\), reaches the maximum \(H(\vc)\), and the second largest NMI, \(\text{NMI}(\vc,2)\), is close to zero.
Therefore, the decrement of \(\text{NMI}(\vc,1)\) implies that the information of one factor diffuses into another latent variable, which we define as ID\@.
The representation can be said to re-entangle in the case of an ID\@.

We monitored \(\text{NMI}(\vc,1)\) and \(\text{NMI}(\vc,2)\) during training with AnnealedVAE on dSprites (training details in Section~\ref{sec:setting}), as shown in Fig.~\ref{fig:NMI_itrs}.
We computed the NMIs for the five factors every 1e4 iterations and presented them in one row.
Ideally, the expanded capacity would promote the model learn new information.
Oppositely, \(\text{NMI}(\vc,1)\) (scale) decreased after 5e4 iterations.
AnnealedVAE suffered ID, which caused the low performance.


\section{Method}

\subsection{Information Freezing}\label{sec:IFP}
\begin{figure}[h]
    \centering
    \includegraphics[width=0.657\linewidth]{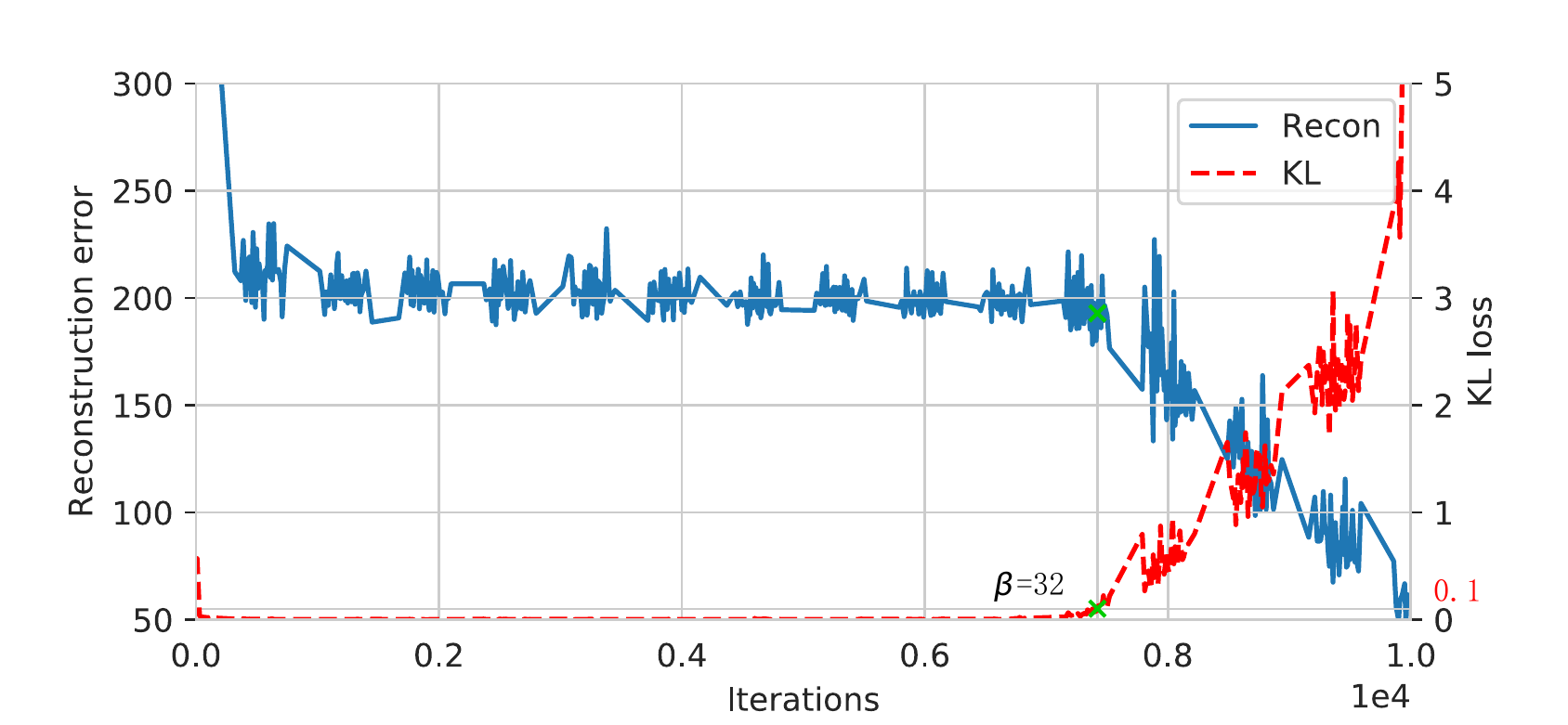}
    \caption{Information freezing. The model starts to learn information at iteration 7500 (\(\beta=32\)), where KL increases and the reconstruction error decreases.}\label{fig:threshold_example}
\end{figure}

\citeauthor{Burgess.2018} proposed that the value of beta in \betavae{} controls the IB between inputs and latent variables, similar to the role of temperature in distillation; a low value of beta encourages the MI \(I(x;z)\), and more information condenses on the latent space.
The IFP is a critical point at which the model starts to learn information from observations.
It is an intrinsic property of a dataset and almost invariant.
Thus, different factors can be identified using IFP.
Intuitively, an approach will work if it progressively disentangles factors when the IFP distributions for these factors are separable.
\begin{definition}
    The IFP is the maximum value of \(\beta\), such that \(I(x;z)>0\) for the \betavae{} objective.
\end{definition}

We introduce the \textbf{annealing test} to determine the IFP for a given dataset.
The objective of the annealing test is the same as that of \betavae, except that it uses an annealing \(\beta\) from a high value to 1 (i.e., it starts with value 200 and ends with value 1).
While the pressure of the KL term decays, there exists a critical point where \(I(x;z)\) increases and the reconstruction error decreases.
For example, we trained the model with an annealing \(\beta\) from 200 to 0 in 100,000 iterations.
As shown in Fig.~\ref{fig:threshold_example}, the IFP is approximately 32 after 7400 iterations.
Roughly, we regard the IFP as the value of beta, where the model learns information (\(I(x;z)\) is over 0.1).

\subsection{DEFT}

\begin{figure}[t]
    \centering
    \subfigure[Forward]{\includegraphics[width=.66\linewidth]{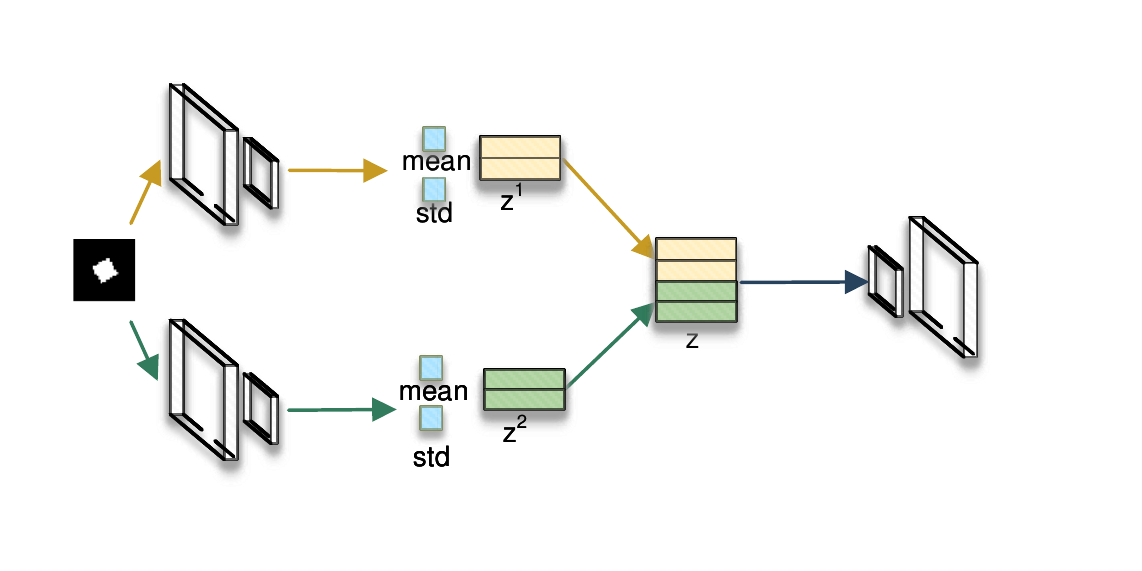}}%
    \subfigure[Backward]{\includegraphics[width=.33\linewidth]{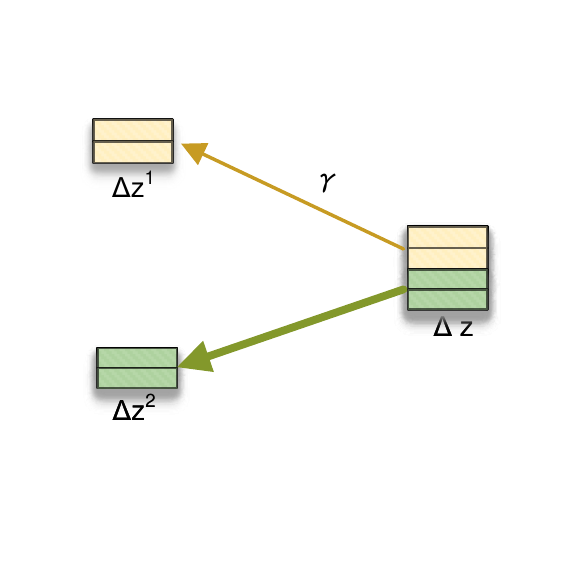}}%
    \caption{Architecture of DEFT\@. Each group had two latent variables. (a) The forward part of DEFT concatenates two groups of latent variables into a four-dimensional vector, and it then feeds to the decoder. (b) The backward part of the DEFT clips the backward information for the old variables $z^1$, \(\Delta z^1=\gamma \Delta z_{1:2},\Delta z^2= \Delta z_{3:4}\).}\label{fig:architecture}
\end{figure}

Inspired by distillation in chemistry, this paper proposes a novel disentanglement approach that is based on \betavae{}, according to IB theory.
It splits the latent variables into G groups, and has K independent encoders for each of these groups.
The decoder takes the concatenation of latent variables of all groups.
DEFT also divides the training process into G stages, such that during the i-th stage, the parameters of all encoders, except the i-th encoder, train with a smaller learning rate that is \(\gamma \) times smaller than the i-th encoder.
In addition, each stage has a decayed coefficient beta of \betavae{}, which controls the IB from inputs to latent variables.
The architecture of the DEFT (\(\text{G}=2, \text{K}=2\)) is shown in Fig.~\ref{fig:architecture}.
The left and right figures show the forward and backward processes of the DEFT\@, respectively.
We show the algorithm of DEFT in Algorithm~\ref{algorithm}, where \(q_{\phi_i}^i(z|x)\) denotes one group of encoders, \(p_\theta(x|z)\) denotes the decoder, and \(\mathcal{L}^1\) denotes the \betavae{} objective.

DEFT chooses a suitable value of beta to separate factors that act as the temperature, such that the desired factor's information passes the bottleneck and freezes into the latent variables.
Each stage disentangles one factor in the ideal situation.
However, the information can be assigned to another variable if the pressure decays (reduces beta), which causes ID\@.
Therefore, backward information scaling is performed for these variables to prevent the diffusion of information into others.

\begin{algorithm}[h]
\SetAlgoLined{}
\begin{algorithmic}
    \STATE {\bfseries Input:} pressure \(\beta_j\), stages G, \(\gamma\)
    \STATE Initialize \(\theta, \left\{ \phi_i \right\}\) for \(p_\theta(x|z), \{q_{\phi_i}^i(z|x)\}\).
    \FOR{\(j=1\) {\bfseries to} G}
        \STATE \(g_\theta, \left\{ g_{\phi_i} \right\} = \nabla_{\theta,\left\{ \phi_i \right\}} \mathcal{L}^1(\theta, \left\{ \phi_i \right\}; \beta_j)\)
        \FOR{\(i=1\) {\bfseries to} G}
            \IF{\(i = j\)}
            \STATE \(\phi_i=\phi_i-g_{\phi_i} \times \mathrm{lr} \)
            \ELSE
            \STATE \(\phi_i=\phi_i-g_{\phi_i} \times \mathrm{lr} \times \gamma\)
            \ENDIF
        \ENDFOR
        \STATE \(\theta=\theta-g_\theta \times \mathrm{lr}\)
    \ENDFOR
\end{algorithmic}
\caption{The algorithm for DEFT.}\label{algorithm}
\end{algorithm}

\section{Experiment}

\subsection{Settings}\label{sec:setting}
In this study, there are two types (standard and lite) of encoders and one decoder architecture, as shown in Table~\ref{tab:architecture}.
One group encoder of DEFT uses the lite architecture and has K latent variables---the dimension of z is \(2\text{K}\) in total; the other approaches use the standard architecture.
All models use the same decoder architecture.
All layers are activated by ReLU\@.
The optimizer is Adam with a learning rate of 5e-4.
The batch size is 256, which accelerates the training process.

\begin{table}[h]
    \caption{Lite encoder, standard encoder, and decoder architecture for all experiments. For dSprites and SmallNORB, \(c=1\). For Color and Scream, \(c=3\).}%
    \label{tab:architecture}
    \centering
    \begin{tabular}{|l|l|l|}
        \toprule
        {Lite Encoder} & {Standard Encoder} & {Decoder} \\
        \midrule
        \(4 \times 4\) conv. Eight Stride 2 & \(4 \times 4\) conv. 32 stride 2 & FC. 256 \\
        \midrule
        \(4 \times 4\) conv. Eight Stride 2 & \(4 \times 4\) conv. 32 stride 2 & FC. \(4 \times 4 \times 64\) \\
        \midrule
        \(4 \times 4\) conv. Sixteen stride 2 & \(4 \times 4\) conv. 64 stride 2 & \(4 \times 4\) upconv. 64 stride 2 \\
        \midrule
        \(4 \times 4\) conv. Sixteen stride 2 & \(4 \times 4\) conv. 64 stride 2 & \(4 \times 4\) upconv. 32 stride 2 \\
        \midrule
        FC. 64.  & FC. 256. & \(4 \times 4\) upconv. 32 stride 2\\
        \midrule
        FC. \(2 \times \text{K}\). & FC. \(2 \times 10\). & \(4 \times 4\) upconv. \(c\). stride 2 \\
        \bottomrule
    \end{tabular}
\end{table}

\subsection{Backward Information Scaling}

\begin{figure}[h!]
    \centering
    \subfigure[\(\text{NMI}(\vc,1)\)]{\includegraphics[width=.5\linewidth]{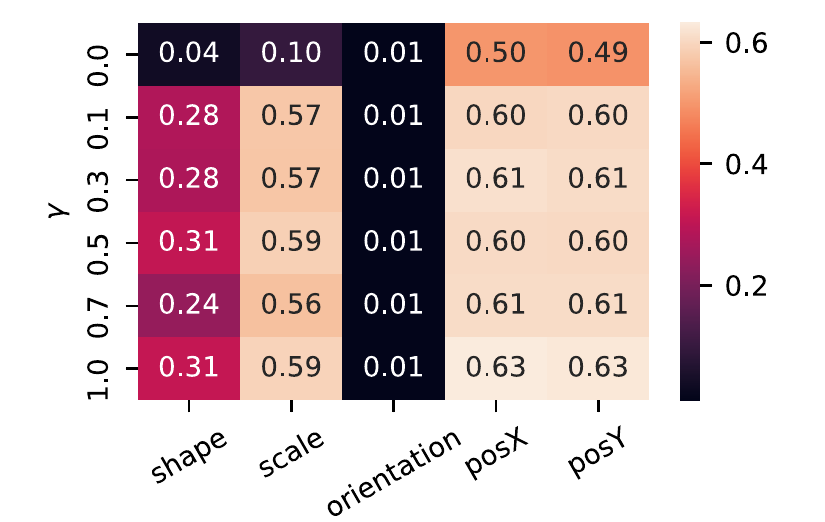}}%
    \subfigure[\(\text{NMI}(\vc,2)\)]{\includegraphics[width=.5\linewidth]{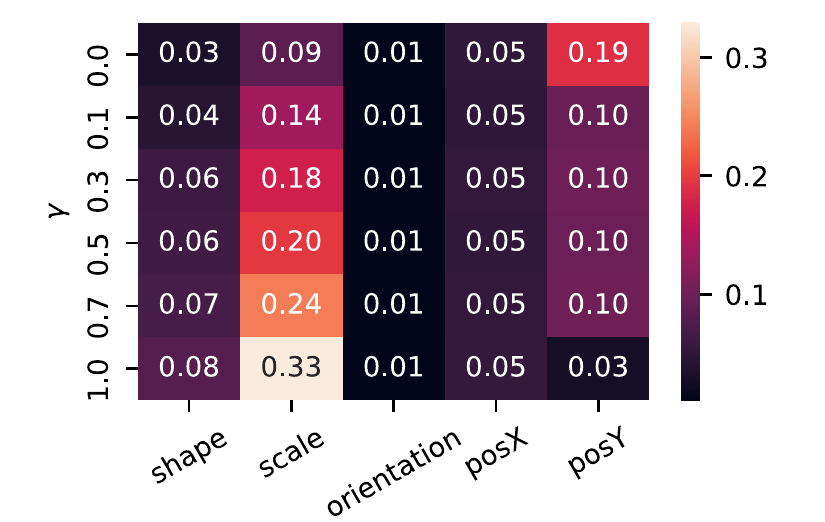}}
    \caption{\(\text{NMI}(\vc,1)\) and \(\text{NMI}(\vc,2)\) for \(\gamma\). Each column shows MI \(I(\vc_i;z)\). A larger value of \(\gamma\) leads to greater ID (b).}\label{fig:gamma}
\end{figure}

\paragraph{Selection of Gamma}
We trained the model in the first stage with \(\beta=70\) and then trained the second stage with \(\beta=30\) for different values of \(\gamma\).
As shown in Fig.~\ref{fig:gamma}, all backward information is clipped when \(\gamma=0\), and \(I(x;z)\) will not increase;
\(\text{NMI}(\vc,2)\) and \(\gamma\) are simultaneously increasing.
In contrast, \(\text{NMI}(\vc,1)\) has a small increment.
A small value of \(\gamma\) is sufficient to learn the majority information, and it also prevents that information from diffusing into another variable.
In practice, \(\gamma=0.1\) achieves a good balance between learning new information and preventing ID.

\paragraph{Comparison}\label{sec:comparison}
AnnealedVAE directly controls the capacity of the latent variables by an annealed increasing parameter, \(C\).
CascadeVAEC increases the capacity by relieving the pressure on the $i$-th latent variable at the $i$-th stage, opening the information flow.
We found that ID is an invisible hurdle for the IB-based approaches, which can be detected by the mean of \(\text{NMI}(\vc_i,1), \text{NMI}(\vc_i,2)\):
\begin{equation}\label{eq:ID}
    \begin{aligned}
        \text{NMI1} =  \sum_{i=1}^{\Vert \vc \Vert} \text{NMI}(\vc_i,1),\\
        \text{NMI2} =  \sum_{i=1}^{\Vert \vc \Vert} \text{NMI}(\vc_i,2).
    \end{aligned}
\end{equation}

\begin{figure}
    \centering
    \subfigure[Differences on controlling capacity.]{\includegraphics[width=\linewidth]{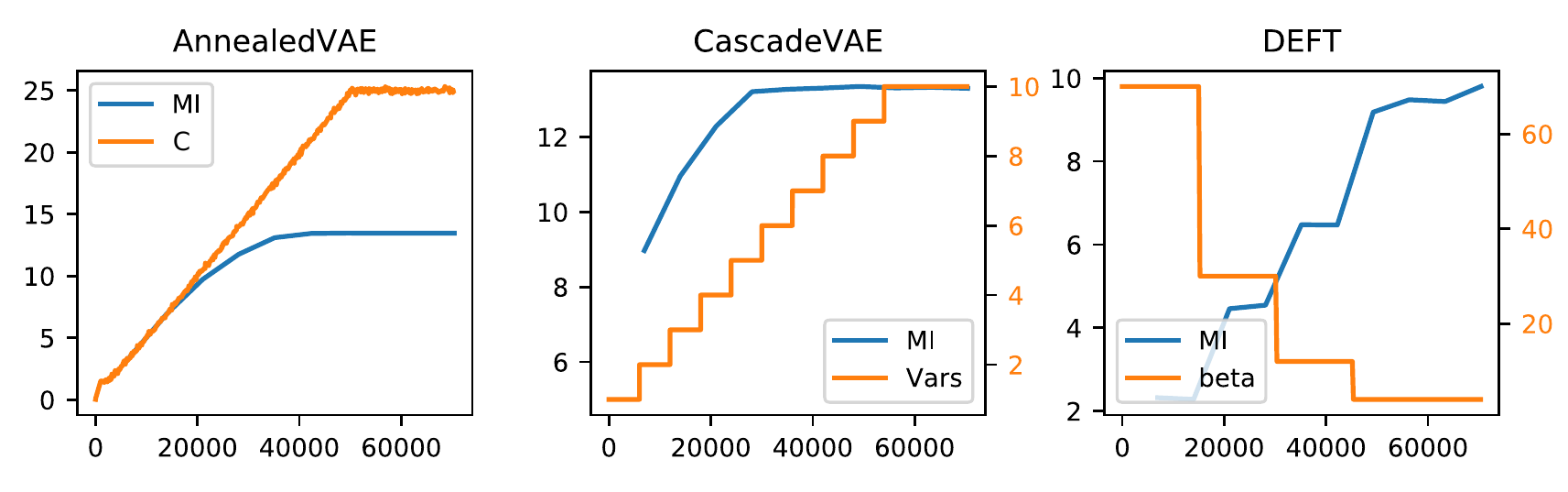}}

    \subfigure[NMI metrics]{\includegraphics[width=\linewidth]{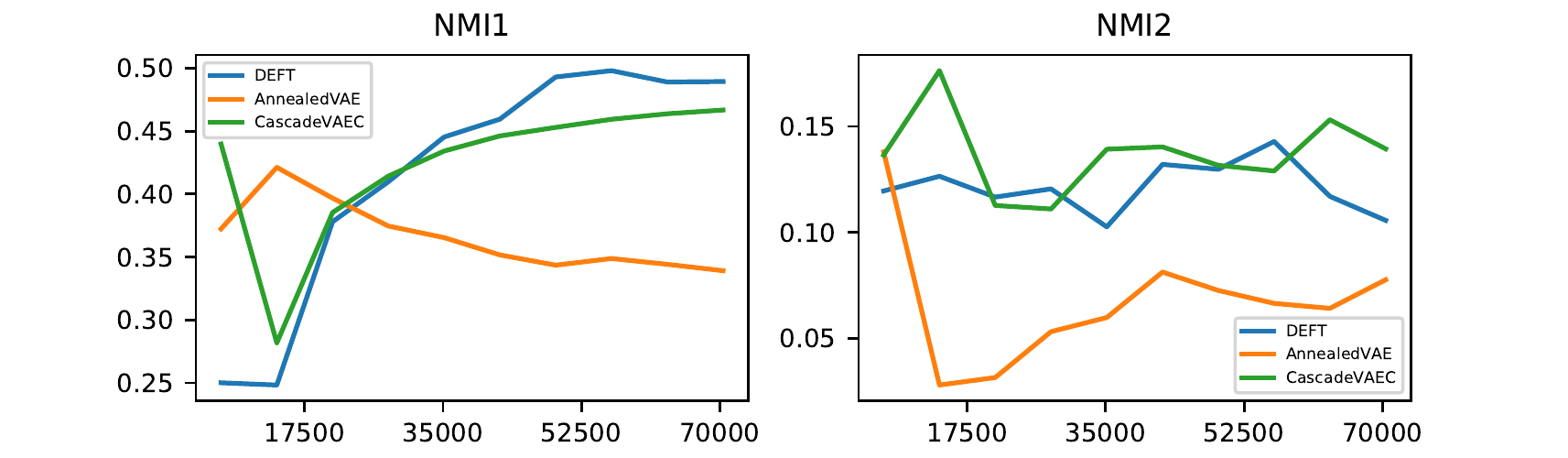}}
    \caption{Comparison of the IB-based approaches. DEFT has a steady increment of MI and NMI1.}\label{fig:comparison}
\end{figure}

We conducted experiments involving AnnealedVAE, CascadeVAEC, and DEFT on dSprites.
As depicted in Fig.~\ref{fig:comparison} (a), AnnealedVAE expands the capacity by a controllable parameter $C$; however, the KL consists of MI, TC, and DWKL.
There is an increasing gap between the MI and $C$. 
At each stage, CascadeVAEC activates a new latent variable to expand the capacity.
DEFT expands the capacity by relieving the pressure.
One can see, DEFT has a steady increment of MI\@.
Both AnnealedVAE and CascadeVAEC suffer from the ID problem (see in Fig.~\ref{fig:comparison} (b)): AnnealedVAE tends to diminish NMI1; CascadeVAEC also has a sharp decrement of NMI1.
Only DEFT remits the ID problem and promotes the disentanglement steadily.

\subsection{Supervised Problem}
\begin{figure}[h]
    \includegraphics[width=\linewidth]{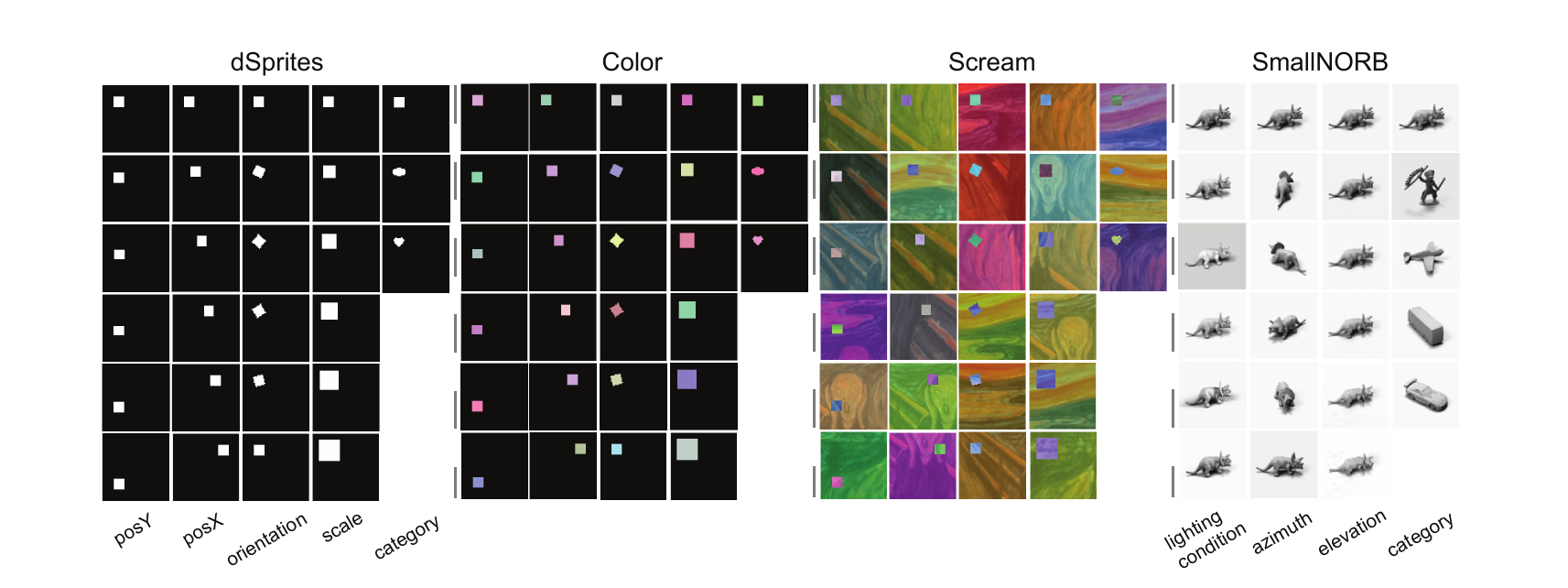}
    \caption{Visualization of datasets. Each row enumerates the image of an explanatory factor. For dSprites, there are five factors: shape, scale, orientation, position X, and position Y. For SmallNORB\@, there are four factors: category, elevation, azimuth, and lighting conditions.}%
    \label{fig:dsprites}
\end{figure}

\paragraph{Dataset Detail}
We compared DEFT with others on dSprites \citep{dsprites17}, color  (color for short), screen  (screen for short), and SmallNORB \citep{DBLP:conf/cvpr/LeCunHB04/smallnorb}.
The images of dSprites are strictly generated by the five factors.
It has three shapes: square, ellipse, and heart; six scale values: 0.5, 0.6, 0.7, 0.8, 0.9, and 1.0, 40 orientation values in [0, 2 pi], 32 position X values, and 32 position Y values.
Two variants of dSprites (color and screen), which introduce random noise, were closer to the true situation.
SmallNORB is generated from 3D objects and is much more complex than 2D shapes.
It contains five generic categories, namely, four-legged animals, human figures, airplanes, trucks, and cars; nine elevation values, i.e., 30, 35, 40, 45, 50, 55, 60, 65, and 70; eighteen azimuth values, 0, 20, 40,\ldots, 340; and six lighting conditions.
Fig.~\ref{fig:dsprites} shows the visualization of the datasets.

\paragraph{Information Freezing Point}
\begin{figure}[h!]
    \centering
    \includegraphics[width=\linewidth]{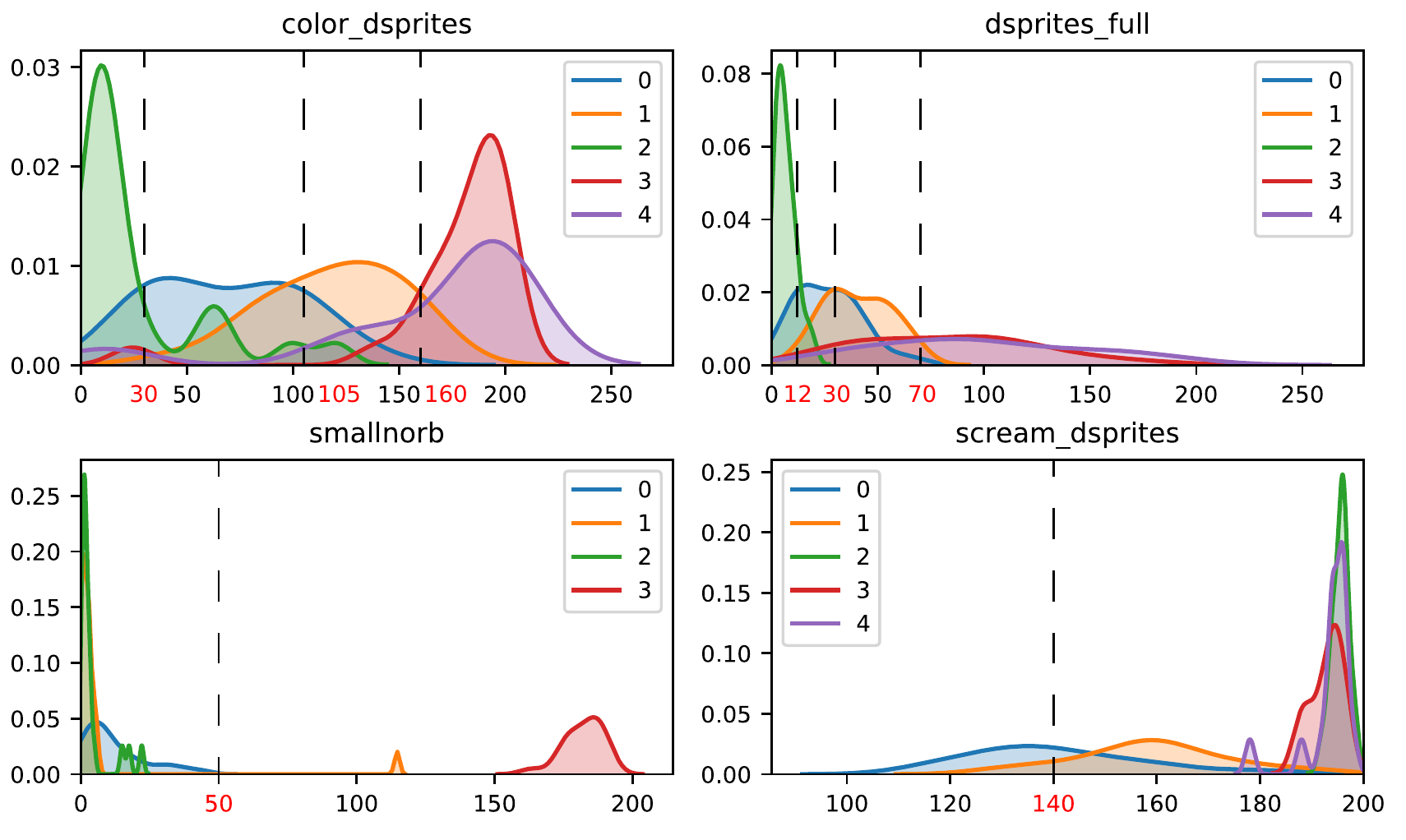}
    \caption{IFP of factors on four datasets. The red number denotes the pressure required to separate these factors. For SmallNORB, 0--3 respectively denote category, elevation, azimuth, and lighting condition. For variants of dSprites, 0--4 respectively denote shape, scale, orientation, position X, and position Y.}\label{fig:thresholds}
\end{figure}

The ideal situation is to find \textit{a set of \(\beta \) to isolate IFPs into several parts without overlaps}.
To obtain the distribution of IFPs with respect to a factor \(\vc_i\), we enumerate all possible values of factor \(\vc_i\) for a random sample, and calculate its ICP using the algorithm introduced in Section~\ref{sec:IFP}.
Then, we repeated the above procedure 50 times to estimate the ICP distribution of \(\vc_i\).

We measured the IFPs of the factors on the four datasets, as shown in Fig.~\ref{fig:thresholds}.
dSprites and Color had more separable IFPs than Scream and SmallNORB\@.
Although the three variants of dSprites have the same factors, their IFPs are different.
The difference in IFP distributions explains why current approaches fail to transfer hyperparameters across different problems in \citet{Locatello.2019}.
Note that the IFP distributions of factors are almost separable for dSprites and Color; the ground-truth factors are independent of the four datasets.
In summary, they are all independent; Scream and SmallNorb are inseparable.
Based on the distribution of IFPs, we summarize the optimal training settings for the DEFT in Table~\ref{tab:deft_setting}.
We tune the hyperparameters of compared approaches with the highest MIG and show theses setting in Table~\ref{tab:others_setting}.

\begin{table}[t!]
    \caption{Experimental settings for DEFT\@. \(\gamma\) is always \(0.1\). The number of epochs is sufficiently large such that the objective converges. The number of latents per encoder (K) is not less than the size of the newly learned factors. The group number (G) is determined by the number of separable areas in Fig.~\ref{fig:thresholds}.}\label{tab:deft_setting}
    \begin{center}
                \begin{tabular}{l|llll}
                    \toprule
                              & G & K & Epochs per Stage       & \(\beta_i\)    \\ \midrule
                    Color     & 4 & 3 & 7     & 160,105,30,4 \\ \midrule
                    dSprites  & 4 & 3 & 7     & 70,30,12,4   \\ \midrule
                    SmallNORB & 2 & 5 & 100   & 50,1        \\ \midrule
                    Scream    & 2 & 5 & 10    & 140,1     \\ 
                    \bottomrule
                \end{tabular}
    \end{center}
\end{table}

\begin{table}[]
    \caption{Experimental settings for compared approaches.}\label{tab:others_setting}
    \centering
    \begin{tabular}{l|llll}
        \toprule
                                          & Color & dSprites & Scream & SmallNORB \\ \midrule
    AnnealedVAE ($C$)                     & 10    & 5        & 25     & 5         \\ \midrule
    \btcvae{} ($\beta$)                   & 10    & 10       & 6      & 1         \\ \midrule
    \betavae{} ($\beta$)                 & 16    & 16       & 6      & 1         \\ \midrule
    CacadeVAE  ($\beta_h$)                & 10    & 10       & 10     & 10  \\
    \bottomrule     
    \end{tabular}
\end{table}

\paragraph{Quantitative Analysis}

\begin{figure*}[h]
    \centering
    \includegraphics[width=\linewidth]{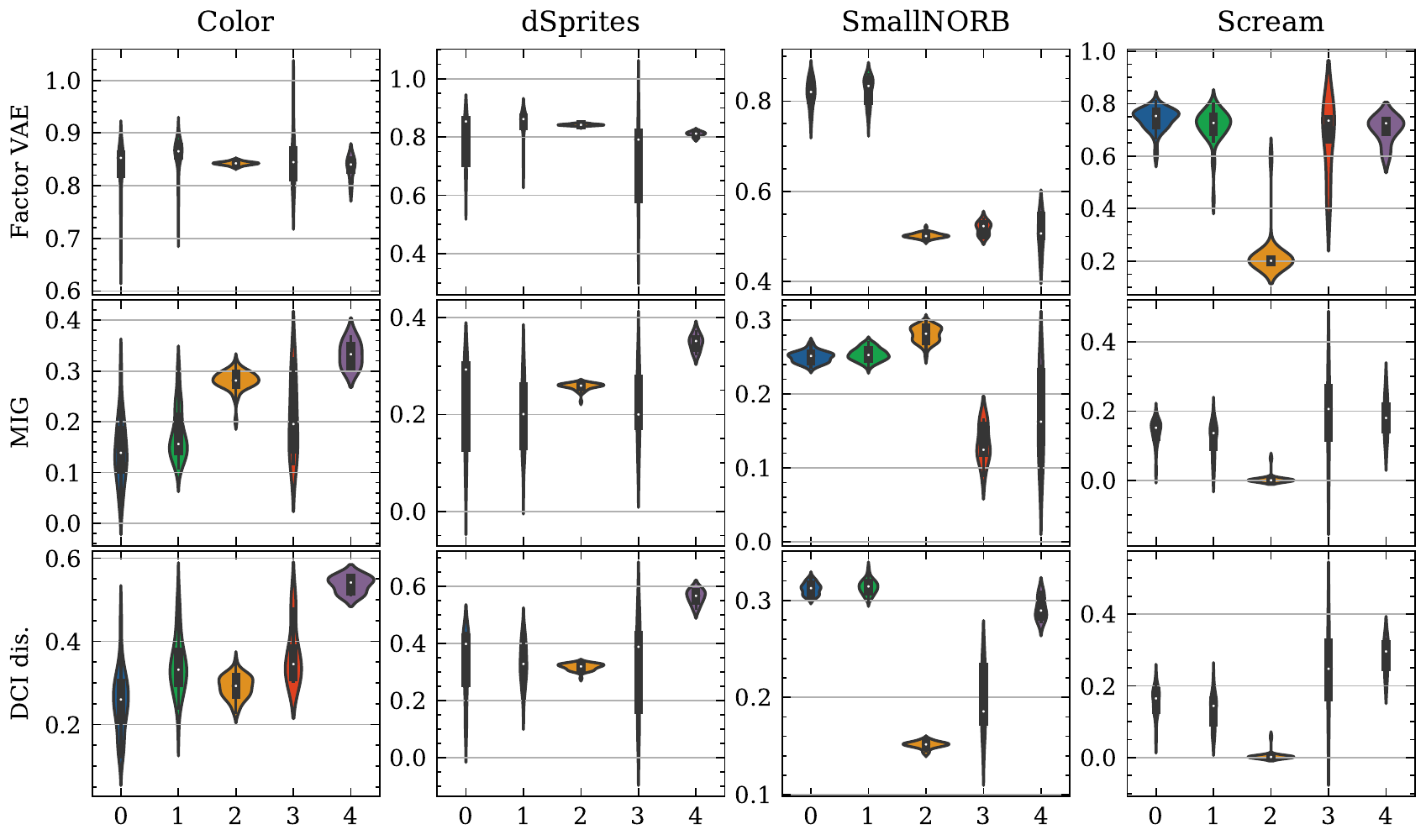}
    \caption{Disentanglement score distribution for different approaches and datasets. Five approaches respectively denote 0=\betavae, 1=\btcvae{}, 2=AnnealedVAE, 3=CascadeVAEC, 4=DEFT.}\label{fig:metrics}
\end{figure*}

\begin{figure}[ht!]
    \centering
    \includegraphics[width=\linewidth]{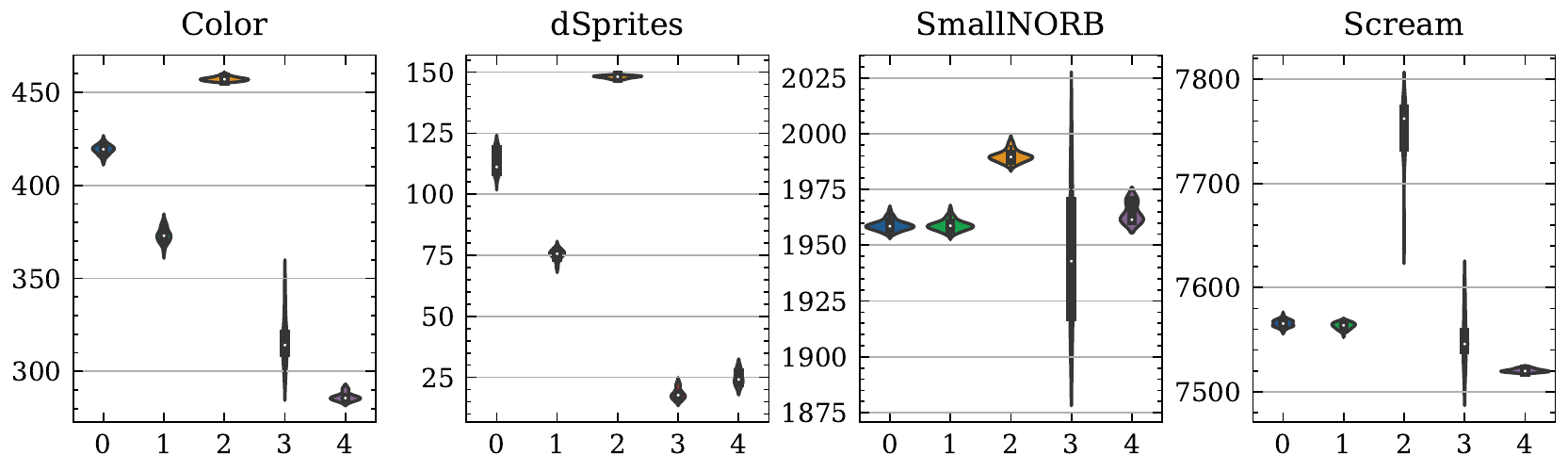}
    \caption{Reconstruction error for different approaches and datasets. Five approaches respectively denote 0=\betavae, 1=\btcvae{}, 2=AnnealedVAE, 3=CascadeVAEC, 4=DEFT.}\label{fig:recon}
\end{figure}

We trained each model 50 times and compared our model with the other four disentanglement approaches on dSprites, Color, Scream, and SmallNORB in Fig.~\ref{fig:metrics}.
All approaches (not just ours) have a lower performance on Scream and SmallNorb (inseparable).
It appears understandable that DEFT fails to handle the situation having inseparable IFP distributions.
However, there is a question as to why these TC-based approaches also failed in an independent but inseparable situation (Scream).
We also show the distributions of the reconstruction error in Fig.~\ref{fig:recon}.
In general, DEFT achieved both a high image quality and disentanglement.

\begin{figure}[ht!]
    \centering
    \subfigure[dSprites]{\includegraphics[width=.5\linewidth]{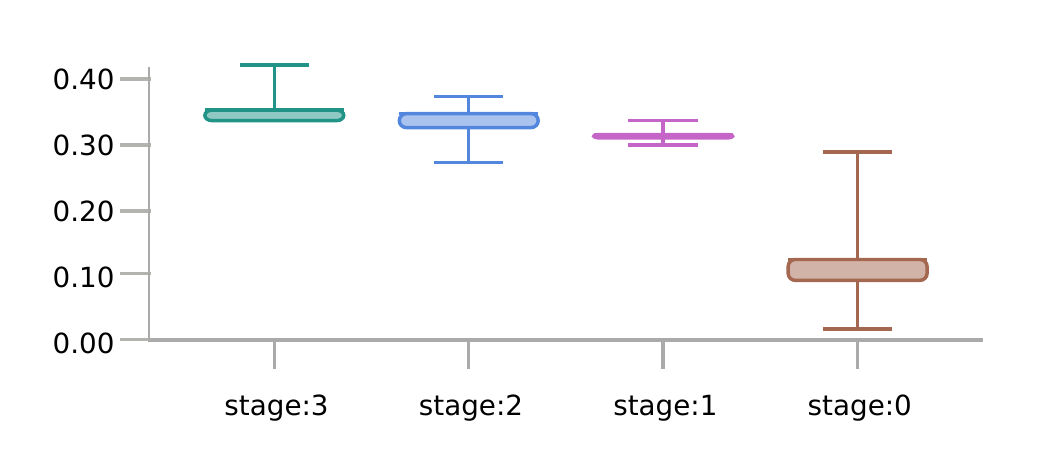}}%
    \subfigure[Color]{\includegraphics[width=.5\linewidth]{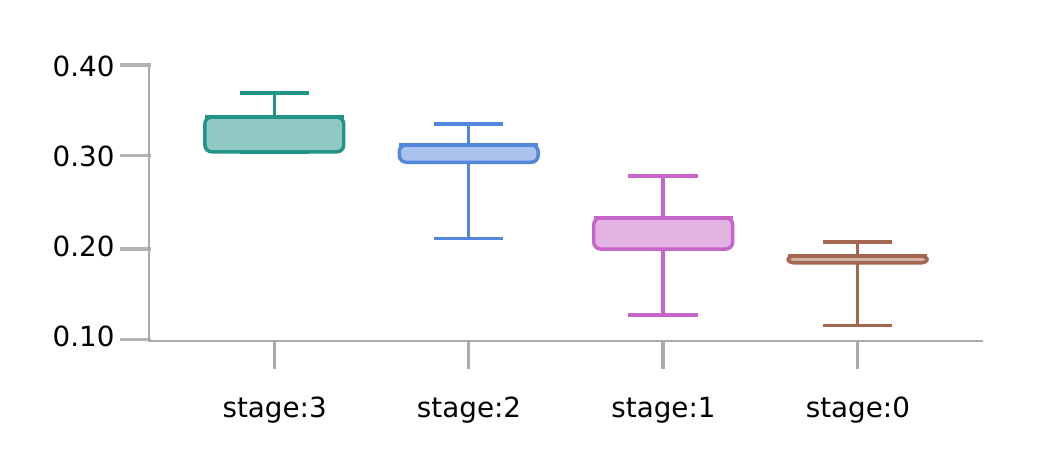}}

    \subfigure[Scream]{\includegraphics[width=.5\linewidth]{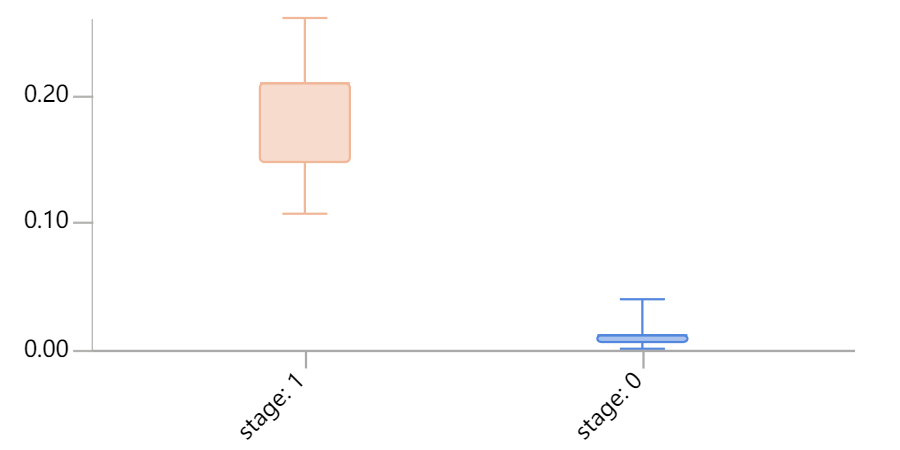}}%
    \subfigure[SmallNORB]{\includegraphics[width=.5\linewidth]{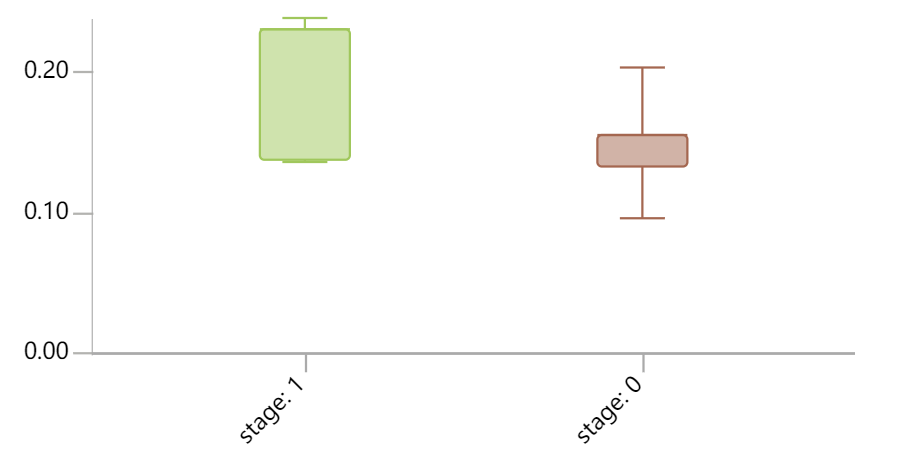}}%
    \caption{MIG distribution of DEFT on four datasets for different stages.}\label{fig:stages}
\end{figure}

The distribution of MIG scores at different stages are shown in Fig.~\ref{fig:stages}.
All experimental results on the four datasets reveal that DEFT obtains low scores in the first stage, and gradually improves disentanglement in the following stages.

\paragraph{Failure rate} 
We define the failure rate as the percentage of models that fail to learn a disentangled representation if the MIG score is lower than 0.1. 
Table~\ref{tab:failure} shows the failure rate. It can be seen that DEFT has the lowest average failure rates. 
Although AnnealedVAE success to disentangle factors on three datasets, it fails to disentangle factors on Scream for most cases.
Note that, it is possible to reduce the failure rate for AnnealedVAE on Scream, but we have tried six settings.
Despite all approaches show low performances on SmallNORB, the failure rates are lower than others. 
From the IFP distributions in Fig.~\ref{fig:thresholds}, we can see that SmallNORB has a separable factor that is easy to be disentangled.
That causes SmallNORB to have a high lower bound of disentanglement but gets a low overall score.
Generally, DEFT significantly decreased the failure rate compared to the other approaches.

\begin{table}[bh]
\caption{Failure rate (\%) for each approach (column) and dataset (row). }\label{tab:failure}
\begin{center}
\begin{small}
\begin{sc}
    \begin{tabular}{lllllll} 
        \toprule
                 &DEFT    & \betavae{} & \btcvae{}  & \text{AnnealedVAE}  & \text{CascadeVAEC}   \\ 
        \midrule
        Color    &0  & 24  & 0 & 0   & 8 \\ 
        dSprites &8  & 16  & 2 & 0  & 0 \\ 
        SmallNORB&0  & 0   & 0  & 0  & 10 \\ 
        Scream   &12 & 12  & 26 & 80 & 25 \\ 
        \bottomrule
    \end{tabular}
\end{sc}
\end{small}
\end{center}
\end{table}

\begin{figure}[h]
    \centering
    \subfigure[Color (0.39)]{\includegraphics[width=.25\linewidth]{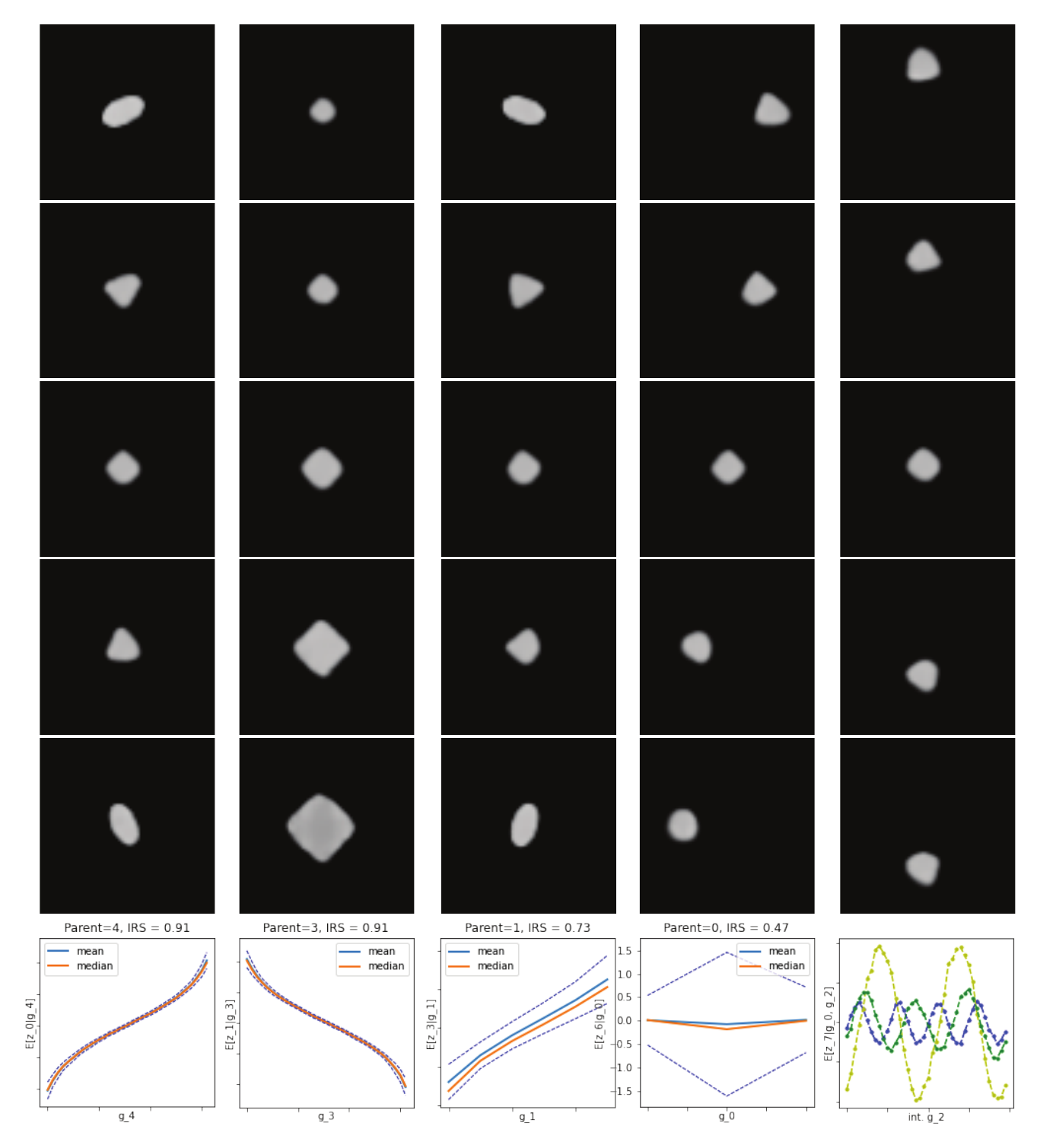}}%
    \subfigure[dSprites (0.41)]{\includegraphics[width=.25\linewidth]{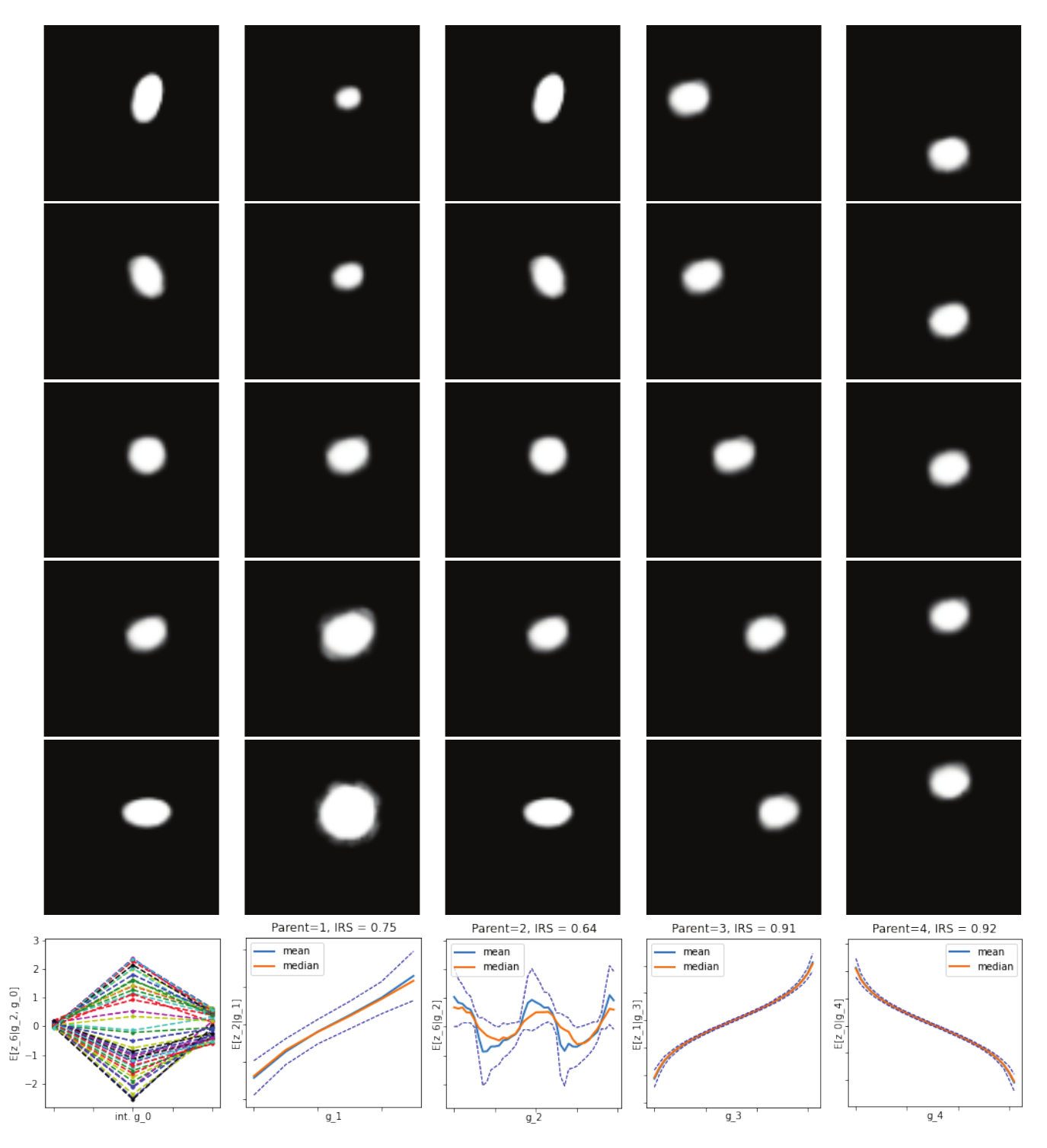}}%
    \subfigure[SmallNORB (0.31)]{\includegraphics[width=.25\linewidth]{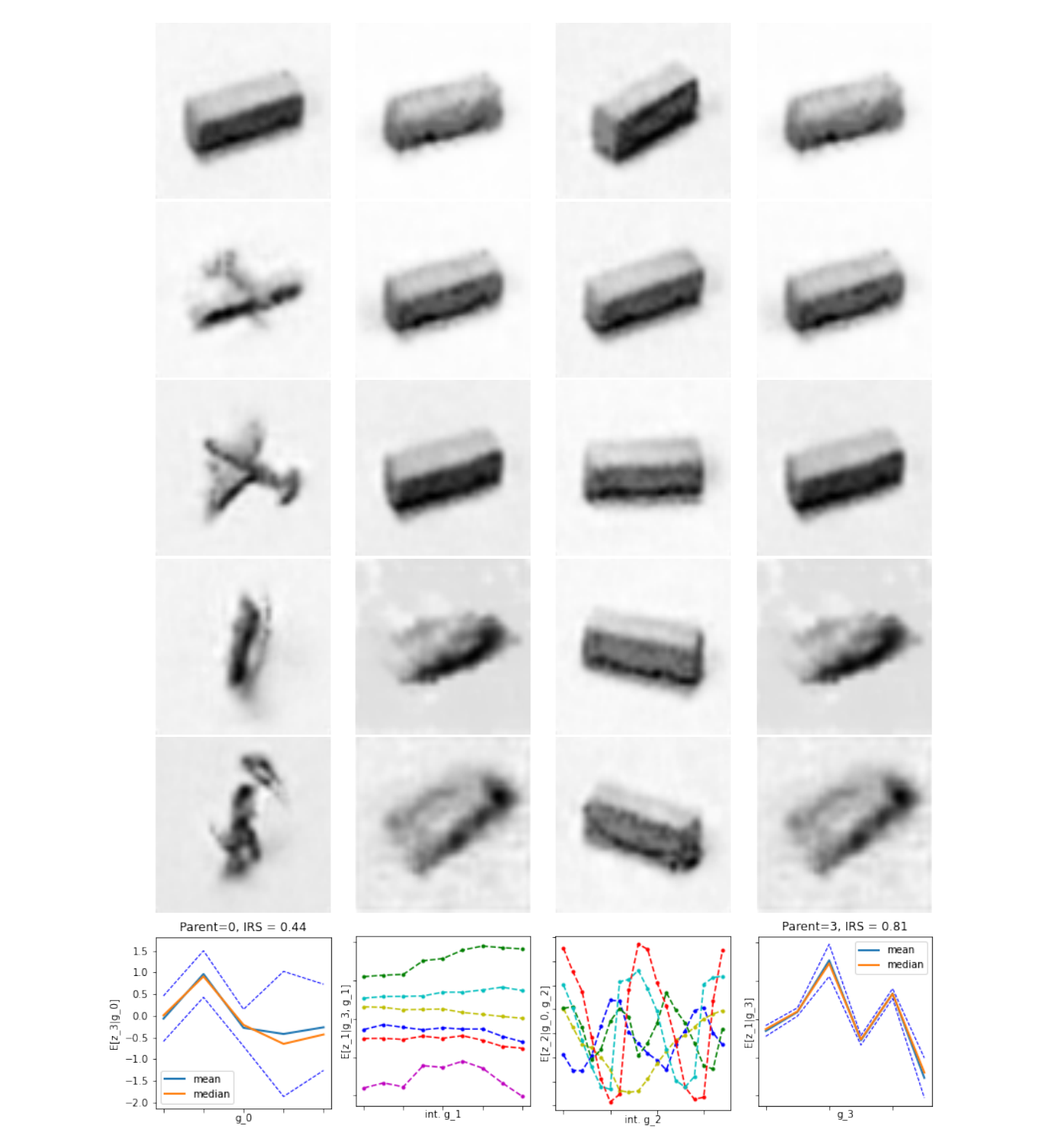}}%
    \subfigure[Scream (0.29)]{\includegraphics[width=.25\linewidth]{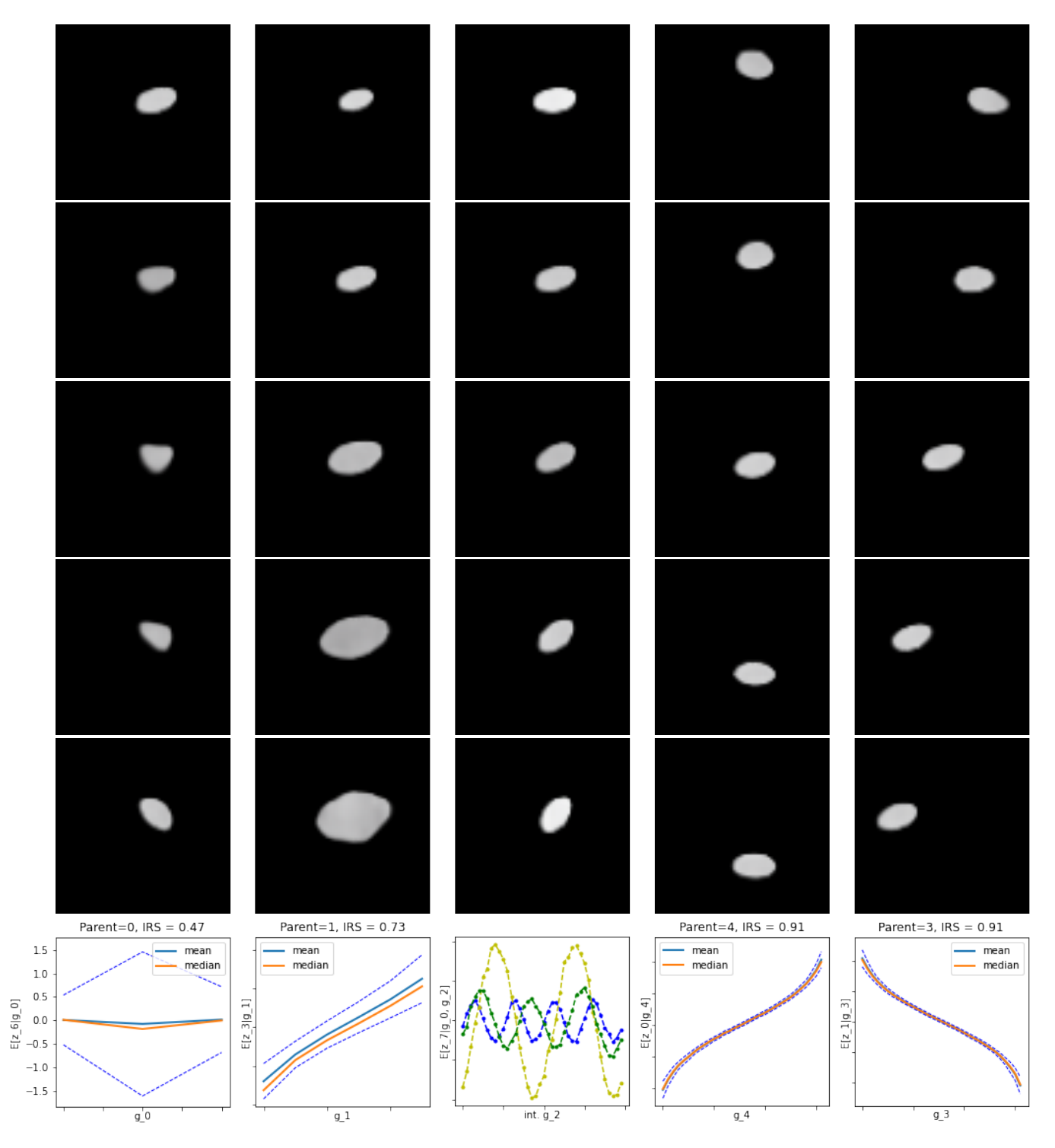}}%
    \caption{Latent traversal of DEFT on four datasets (MIG score). Each column shows the images of traversing a latent \(z_i\) and its VIR (last row) \citep{suter2019robustly}.}\label{fig:traversal}
\end{figure}

\paragraph{Qualitative Analysis}
\citet{Higgins.2017} introduced the latent traversal to visualize the generated images through the traversal of a single latent \(z_i\).
Fig.~\ref{fig:traversal} shows the latent traversal of the best model with the highest MIG score.
One can see the intrinsic relationship between IFP and disentanglement.
Orientation has the lowerest IFP among all factors; meanwhile, it is the hardest one to be disentangled for all approaches.
For SmallNORB, the lighting condition (3) is separable with others, which is easy to be disentnagled. 
For Scream, three factors have similar IFP distributions, and it is also a hard problem for the disentanglement approaches.

\subsection{Correlative but Separable}
\begin{figure}
    \centering
    \subfigure[dataset]{\includegraphics[width=.2\linewidth]{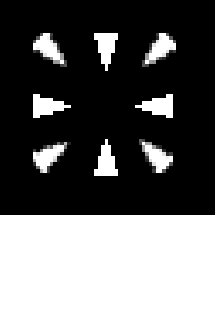}}\hskip 0.1in%
    \subfigure[NMI]{\includegraphics[width=.6\linewidth]{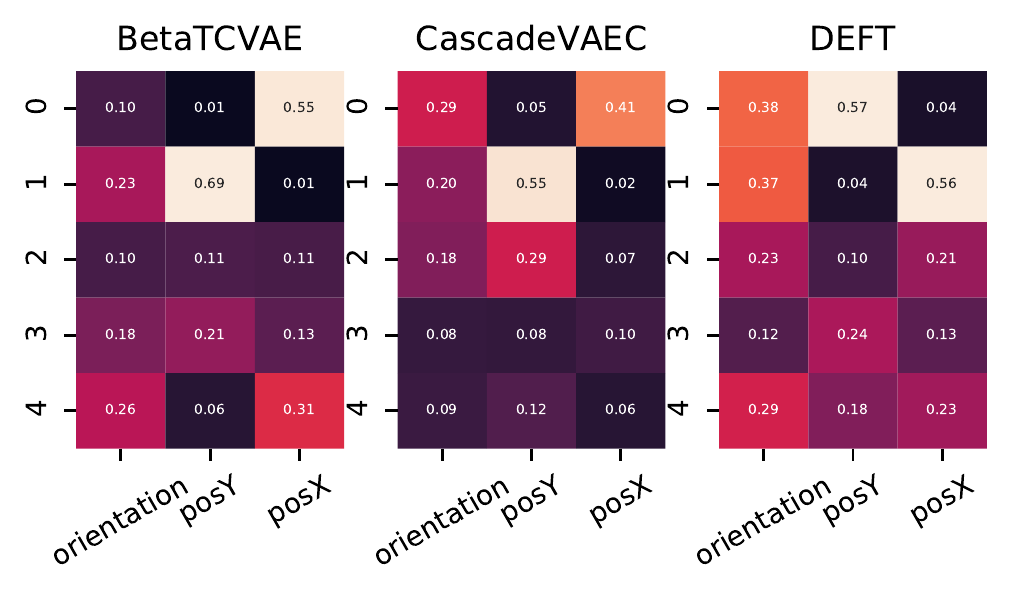}}

    \subfigure[Disentanglement score distribution for different approaches.]{\includegraphics[width=\linewidth]{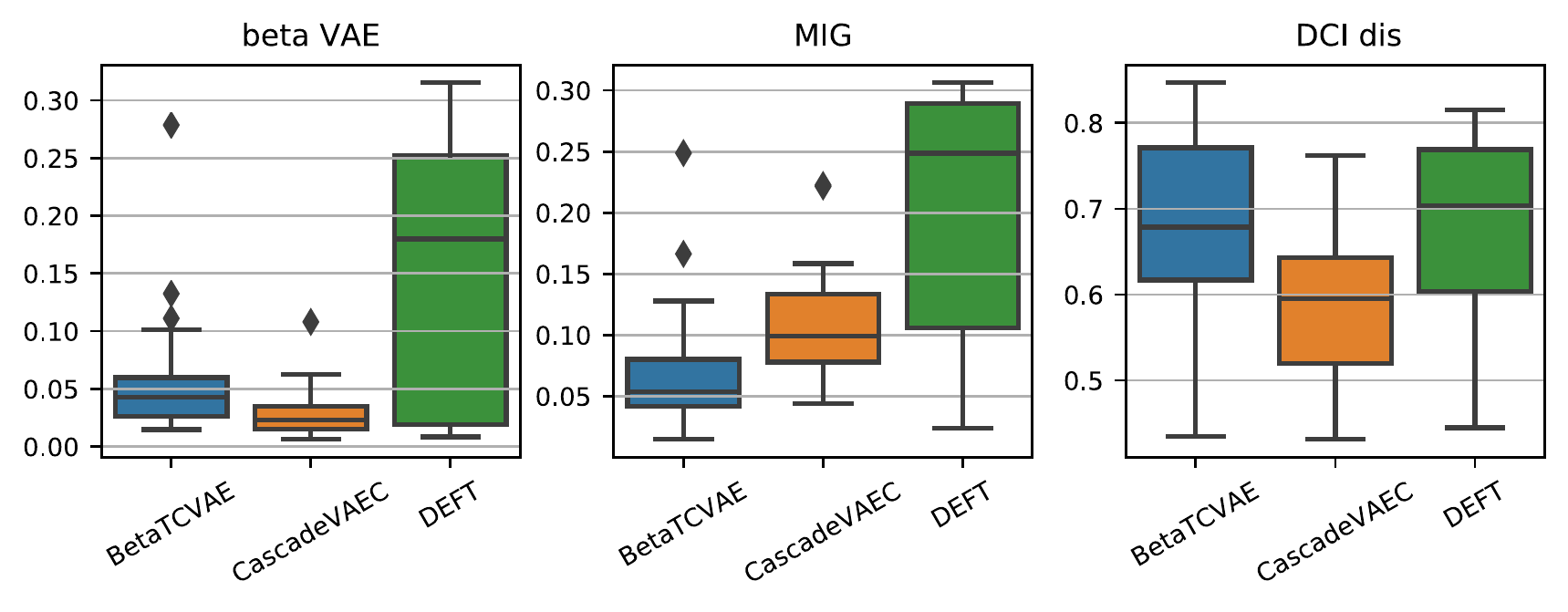}}
    \caption{(a) Dataset visualization. (b) NMI matrix \(I(z_i;c_j)\) for three approaches. (c) Disentanglement scores for different approaches.}\label{fig:correlation}
\end{figure}

\begin{figure}
    \centering
    \subfigure[\btcvae{}]{\includegraphics[width=.33\linewidth]{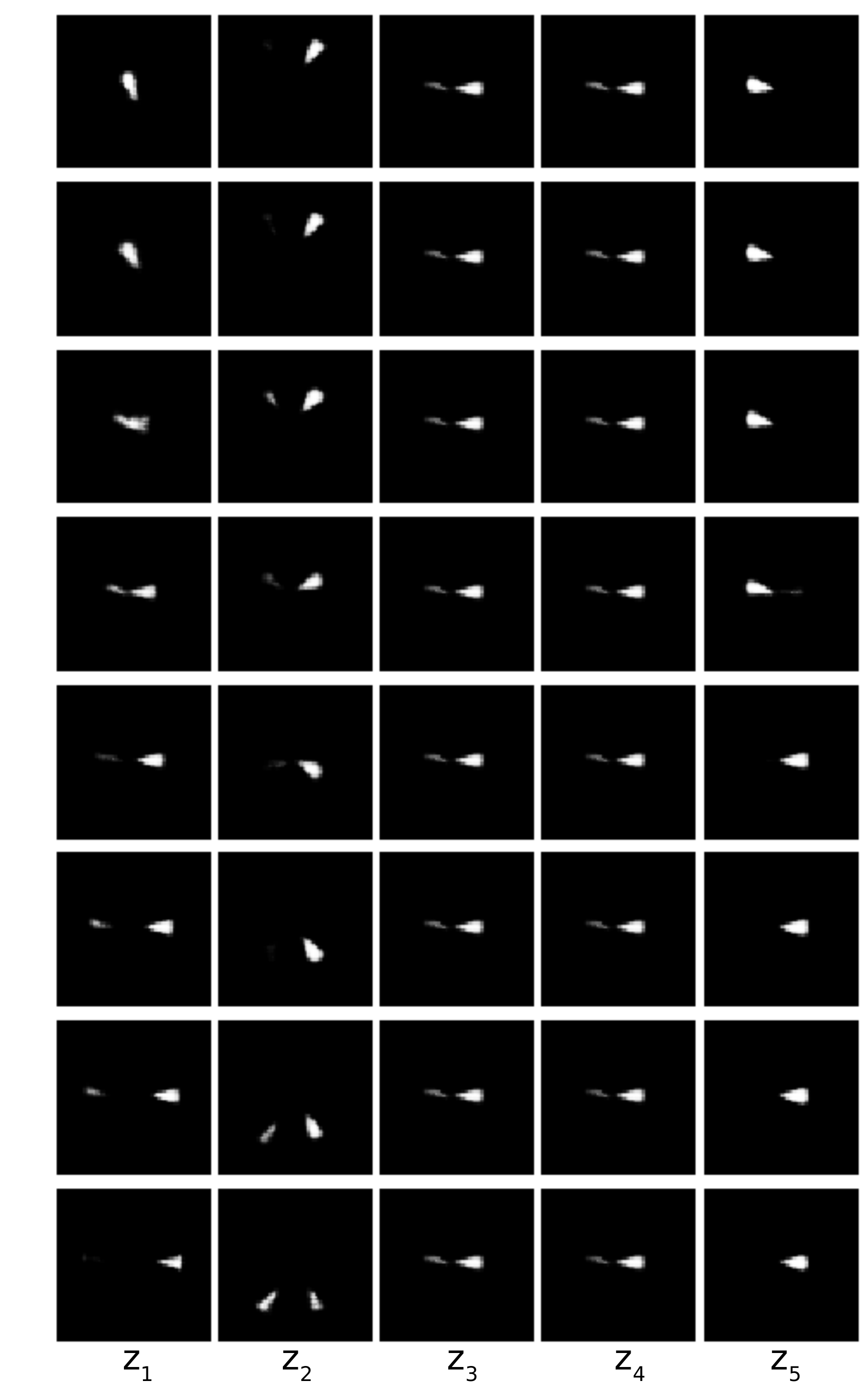}}%
    \subfigure[CascadeVAEC]{\includegraphics[width=.33\linewidth]{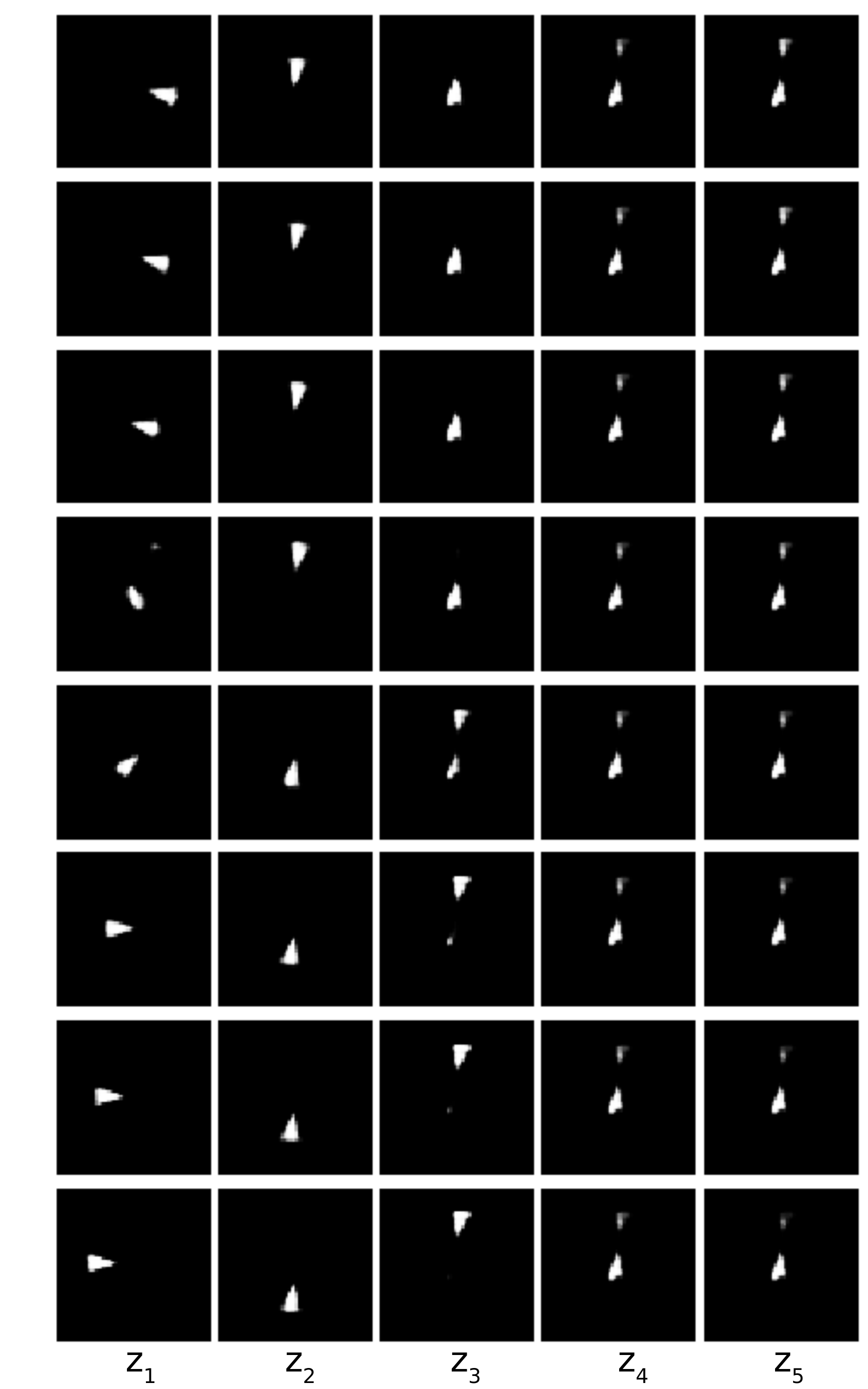}}%
    \subfigure[DEFT]{\includegraphics[width=.33\linewidth]{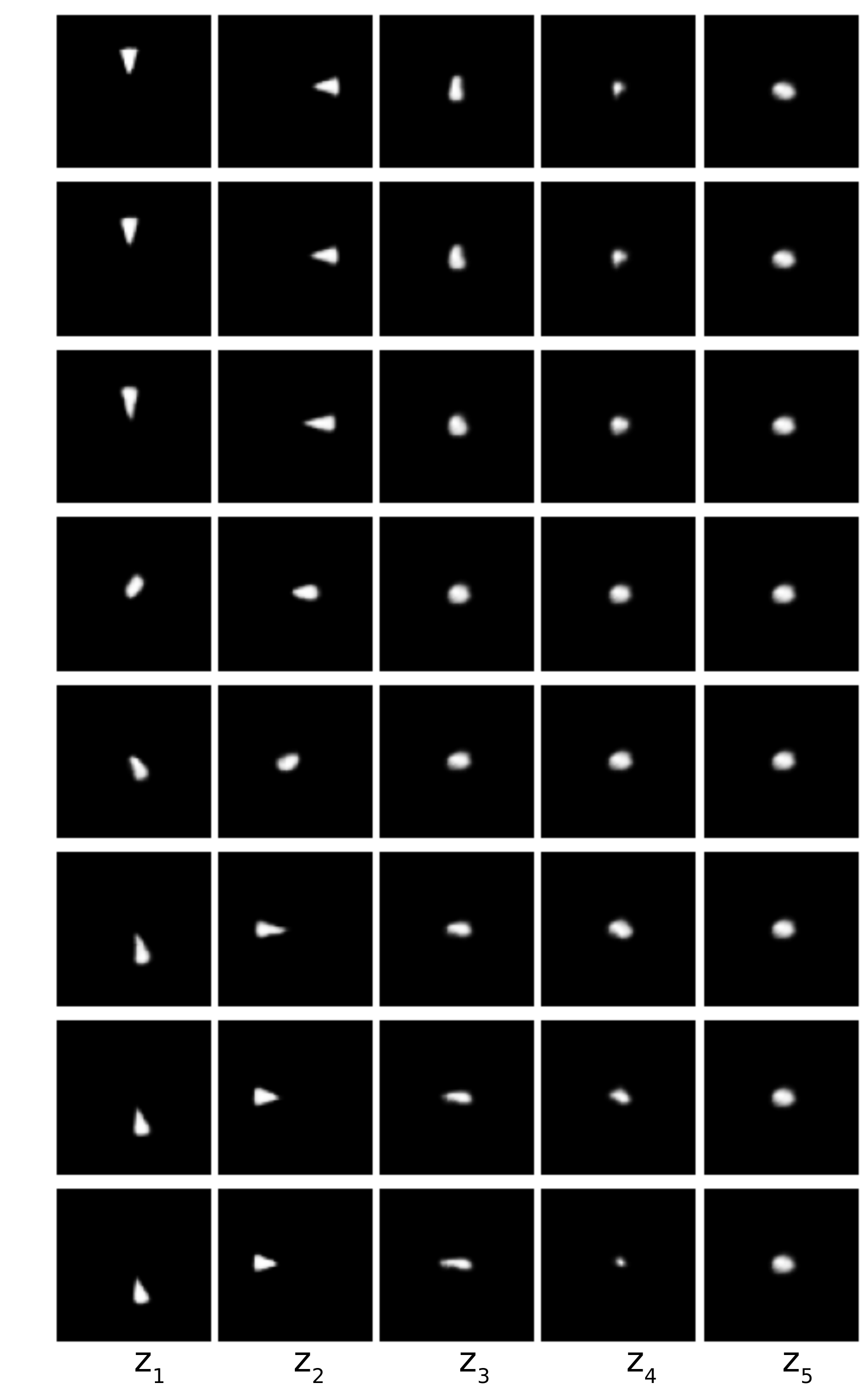}}
    \caption{Latent traversal of CascadeVAEC, \btcvae{}, and DEFT on the separable but correlative dataset.}%
    \label{fig:correlation_traversal}
\end{figure}

To demonstrate the superiority of the IB approaches, we built a dataset of a triangle with three factors (posX, posY, and orientation). 
PosX and posY were independent; however, the orientation pointed to the center of the canvas \(\theta = \arctan(\text{posY}-16,\text{posX}-16)\), as shown in Fig.~\ref{fig:correlation}~(a).
We trained CascadeVAEC, \btcvae{} ($\beta=6$), and DEFT ($\text{K}=2,\text{G}=2$) within 10,000 steps and repeated 10 times.
Fig.~\ref{fig:correlation}~(a) illustrates this toy dataset. 
Each sample contains one triangle which has a unique position and points to the center. 
From Fig.~\ref{fig:correlation}~(b), all three approaches disentangle posX and posY successfully.
However, only DEFT extracts orientation information ($I(z_4;\text{orientation})$ is high, $I(z_4;\text{poxX})$ and $I(z_4;\text{poxY})$ are low).
DEFT has higher disentanglement scores for all three metrics, as shown in Fig.~\ref{fig:correlation}~(c).
The latent traversal in Fig.~\ref{fig:correlation_traversal} shows that DEFT has a high image quality and separated orientation information.
The correlation made it difficult for \betavae{} to disentangle orientation.

\subsection{Unsupervised Problem}
3D Chairs \citep{Aubry2014Chairs} is an unlabeled dataset containing 1394 3D models from the Internet.

\begin{figure}
    \subfigure[dSprites]{\includegraphics[width=.5\linewidth]{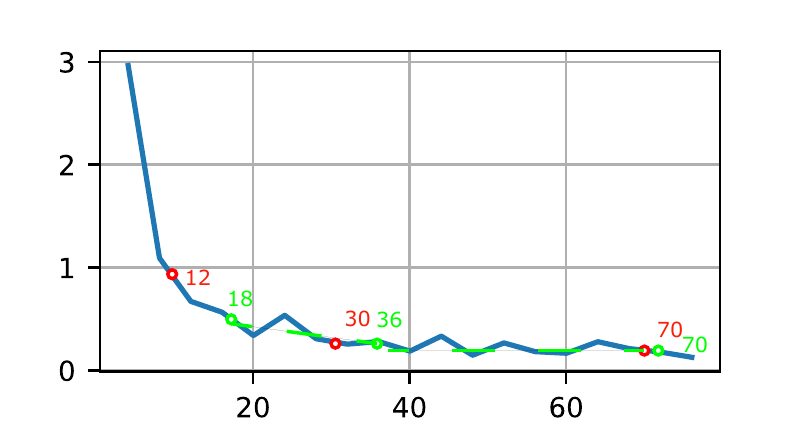}}%
    \subfigure[3D Chairs]{\includegraphics[width=.5\linewidth]{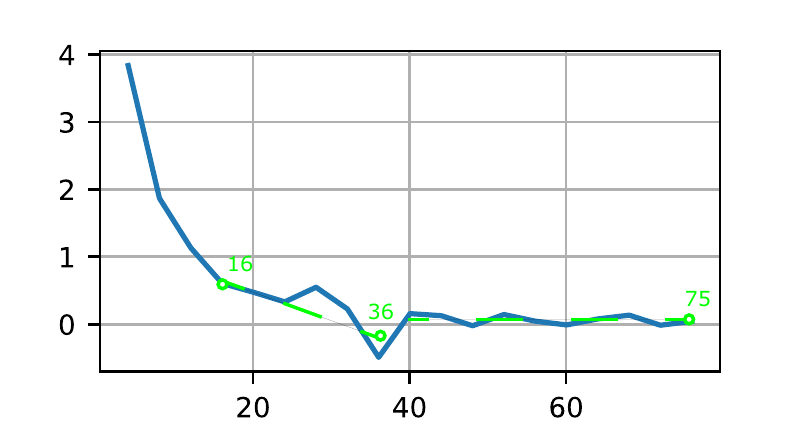}}
    \caption{Information increment variability. The red point denotes the selective separation of the IFP distributions. The green point is the mutation point of the inflation increment. The green broken line denotes the tendency of the growth increment.}\label{fig:anneal}
\end{figure}

\begin{figure}[t]
    \centering
    \includegraphics[width=\linewidth]{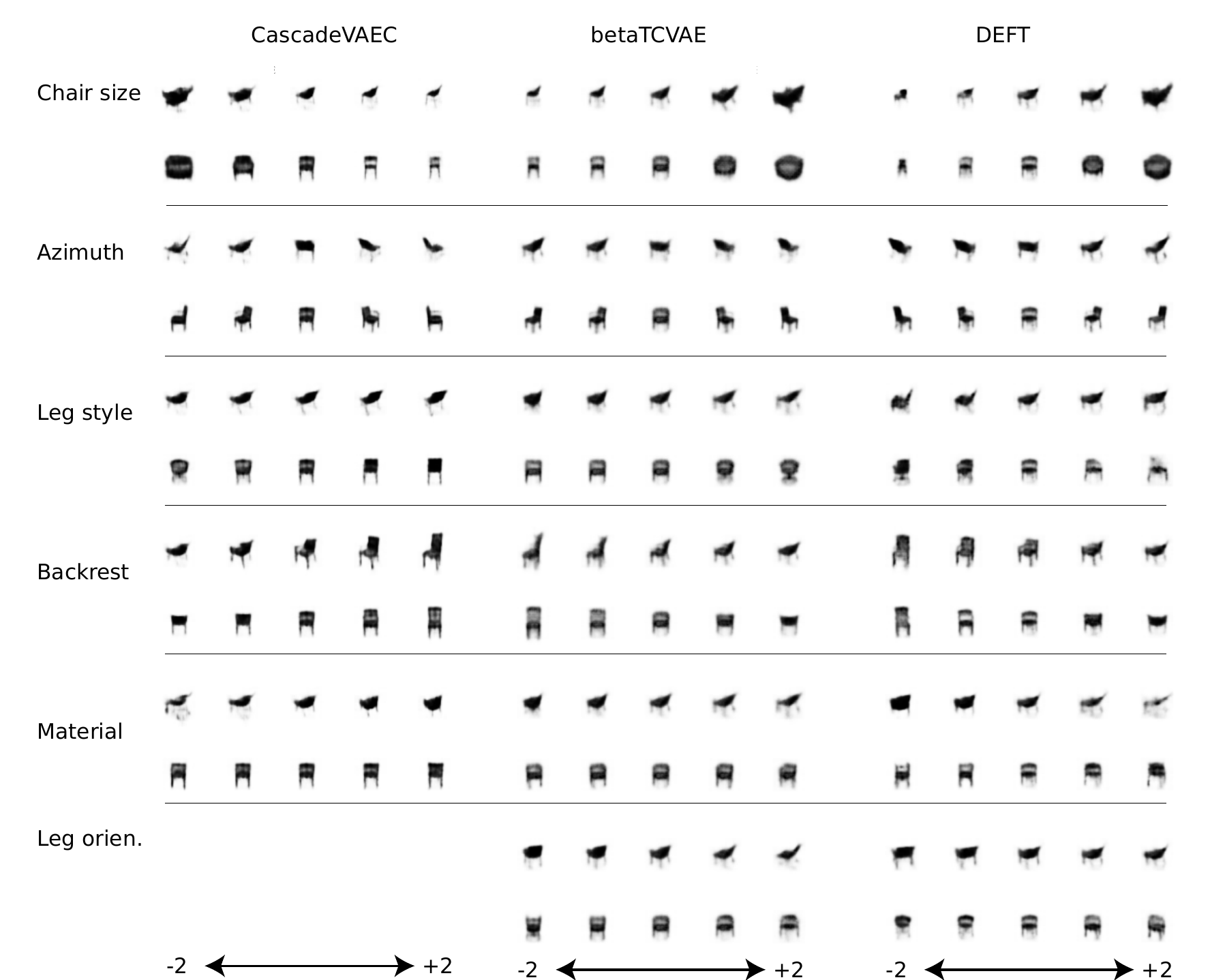}
    \caption{Latent traversal on 3D Chairs. The latent traverses from -2 to 2.}\label{fig:chairs_traversal}
\end{figure}

\paragraph{Annealing Test without Supervision}
The label information is unavailable for the common situations.
Therefore, the factor's IFP distribution is hard to be obtained.
Alternatively, we calculate the upper bound of IFP distribution for the unsupervision setting.
Intuitively, the rate of information increment changes if there is a new factor starting to freeze.
We conducted an annealing test on dSprites and 3D chairs without labels and plotted the curves of beta vs.\ \(\Delta I(x;z)\) in Fig.~\ref{fig:anneal}.
This method is in agreement with the upper bound of the IFP distribution for position and scaling, as shown in Fig.~\ref{fig:anneal} (a).
One can recognize four points where the latent information suddenly increases: 36 and 16 from Fig.~\ref{fig:anneal} (b).
Though this method needs human participation, we only show the potency to develop a fully unsupervised procedure for the separations.
Therefore, we set \(\text{K=3},\text{G=3}, \beta_j=\{36, 16, 1\}\) for 3D Chairs and trained the DEFT 20 epochs per stage.
We compared the performance with \btcvae{} and CascadeVAEC on 3D Chairs, as shown in Fig.~\ref{fig:chairs_traversal}.
We notice that DEFT can learn one additional interpretable property compared with CascadeVAEC: leg orientation.



\section{Conclusion}
Based on existing studies involving IBs, we have developed new insights into the reason for which these approaches have lower performances than the TC-based. 
In particular, we identified the IFP distributions for each factor by performing an annealing test, and a dataset was easily disentangled if the IFP distributions were separable.
Furthermore, we found that the ID problem is an invisible hurdle that prevents steady improvements in disentanglement.
We proposed DEFT to retain the learned information by scaling the backward information. 
To do this, we proposed a learning rate decay \(\gamma\) on these encoders, rather than delivering backward information equally.
Our results show that approaches that are based on IBs are competitive and have the potential to solve problems with correlative factors.

We varified the ID problem that causes the low performance of IB-based approaches. However, as a plain solution, the DEFT method still needs to be further improved. In the future, an automatic way to adjust the best separation of IFP distribution is highly required.

\section*{Acknowledgement}
This work was supported by National Natural Science Foundation of China under Grant No. 61872419,  No. 62072213, No. 61873324. Taishan Scholars Program of Shandong Province, China, under Grant No. tsqn201812077.